\documentclass[journal]{IEEEtran}
\usepackage{cite}
\ifCLASSINFOpdf
  \usepackage[pdftex]{graphicx}
  \graphicspath{{figures/}}
\else
\fi
\usepackage{amsmath}
\usepackage{amssymb}
\usepackage{algorithm}
\usepackage{algpseudocode}

\algnewcommand\algorithmicforeach{\textbf{for each}}
\algdef{S}[FOR]{ForEach}[1]{\algorithmicforeach\ #1\ \algorithmicdo}

\algrenewcommand\algorithmicindent{1.0em}

\ifCLASSOPTIONcompsoc
  \usepackage[caption=false,font=normalsize,labelfont=sf,textfont=sf]{subfig}
\else
  \usepackage[caption=false,font=footnotesize]{subfig}
\fi
\usepackage{booktabs,makecell}
\usepackage{siunitx}

\usepackage{multirow}
\usepackage{hhline}

\usepackage{url}
\makeatletter
\let\NAT@parse\undefined
\makeatother
\usepackage{hyperref}  %hyperref still needs to be put at the end!

\usepackage{array}
\newcolumntype{L}[1]{>{\raggedright\arraybackslash}p{#1}}
\newcolumntype{C}[1]{>{\centering\arraybackslash}p{#1}}
\newcolumntype{R}[1]{>{\raggedleft\arraybackslash}p{#1}}

\usepackage{rotating,siunitx}

\usepackage{color,soul}
\sethlcolor{yellow}

\soulregister\cite7
\soulregister\citep7
\soulregister\citet7
\soulregister\ref7
\soulregister\pageref7

\begin{document}
%
% paper title
% Titles are generally capitalized except for words such as a, an, and, as,
% at, but, by, for, in, nor, of, on, or, the, to and up, which are usually
% not capitalized unless they are the first or last word of the title.
% Linebreaks \\ can be used within to get better formatting as desired.
% Do not put math or special symbols in the title.
\title{3D Reconstruction from Spherical Images: A Review of Techniques, Applications, and Prospects}
%
%
% author names and IEEE memberships
% note positions of commas and nonbreaking spaces ( ~ ) LaTeX will not break
% a structure at a ~ so this keeps an author's name from being broken across
% two lines.
% use \thanks{} to gain access to the first footnote area
% a separate \thanks must be used for each paragraph as LaTeX2e's \thanks
% was not built to handle multiple paragraphs
%

\author{San~Jiang,
	    Yaxin~Li,
	    Duojie~Weng,
	    Kan~You,
        Wu~Chen% <-this % stops a space
\thanks{S. Jiang and K. You are with School of Computer Science, China University of Geosciences, Wuhan 430074, China; S. Jiang is also with Department of Land Surveying and Geo-Informatics, The Hong Kong Polytechnic University, Hong Kong 999077, China.
E-mail: \textit{jiangsan}@cug.edu.cn, \textit{youkan}@cug.edu.cn.}
\thanks{Y. Li, D. Weng, and W. Chen are with Department of Land Surveying and Geo-Informatics, The Hong Kong Polytechnic University, Hong Kong 999077, China.
E-mail: \textit{yaxin.pu.li}@connect.polyu.hk,
\textit{ceweng}@polyu.edu.hk, \textit{wu.chen}@polyu.edu.hk.
\textit{(Corresponding author: Wu Chen)}}}

\maketitle

% As a general rule, do not put math, special symbols or citations
% in the abstract or keywords.
\begin{abstract}
3D reconstruction plays an increasingly important role in modern photogrammetric systems. Conventional satellite or aerial-based remote sensing (RS) platforms can provide the necessary data sources for the 3D reconstruction of large-scale landforms and cities. Even with low-altitude UAVs (Unmanned Aerial Vehicles), 3D reconstruction in complicated situations, such as urban canyons and indoor scenes, is challenging due to frequent tracking failures between camera frames and high data collection costs. Recently, spherical images have been extensively used due to the capability of recording surrounding environments from one camera exposure. In contrast to perspective images with limited FOV (Field of View), spherical images can cover the whole scene with full horizontal and vertical FOV and facilitate camera tracking and data acquisition in these complex scenes. With the rapid evolution and extensive use of professional and consumer-grade spherical cameras, spherical images show great potential for the 3D modeling of urban and indoor scenes. Classical 3D reconstruction pipelines, however, cannot be directly used for spherical images. Besides, there exist few software packages that are designed for the 3D reconstruction from spherical images. As a result, this research provides a thorough survey of the state-of-the-art for 3D reconstruction from spherical images in terms of data acquisition, feature detection and matching, image orientation, and dense matching as well as presenting promising applications and discussing potential prospects. We anticipate that this study offers insightful clues to direct future research.
\end{abstract}

% Note that keywords are not normally used for peerreview papers.

% For peer review papers, you can put extra information on the cover
% page as needed:
% \ifCLASSOPTIONpeerreview
% \begin{center} \bfseries EDICS Category: 3-BBND \end{center}
% \fi
%
% For peerreview papers, this IEEEtran command inserts a page break and
% creates the second title. It will be ignored for other modes.
\IEEEpeerreviewmaketitle

\section{Introduction}
\label{sec:1}
% The very first letter is a 2 line initial drop letter followed
% by the rest of the first word in caps.
% 
% form to use if the first word consists of a single letter:
% \IEEEPARstart{A}{demo} file is ....
% 
% form to use if you need the single drop letter followed by
% normal text (unknown if ever used by the IEEE):
% \IEEEPARstart{A}{}demo file is ....
% 
% Some journals put the first two words in caps:
% \IEEEPARstart{T}{his demo} file is ....
% 
% Here we have the typical use of a "T" for an initial drop letter
% and "HIS" in caps to complete the first word.
%\IEEEPARstart{T}{his} demo file is intended to serve as a ``starter file''
%for IEEE journal papers produced under \LaTeX\ using
%IEEEtran.cls version 1.8b and later.
%% You must have at least 2 lines in the paragraph with the drop letter
%% (should never be an issue)
%I wish you the best of success.

\IEEEPARstart{3}{D} reconstruction is an increasingly critical module in recent photogrammetric systems. It has been extensively utilized for constructing digital cities\cite{xiong2015flexible}, documenting cultural heritages\cite{murtiyoso2017documentation}, and inspecting tunnel cracks\cite{liao2022automatic}, etc. 3D reconstruction can be implemented by using varying instruments, e.g., LiDAR (Light Detection and Ranging) scanners, TOF (Time of Flight) sensors, and optical cameras. The popularity of image sensors and the development of processing techniques have led to the vast usage of image-based 3D reconstruction techniques among all available sensors in the field of photogrammetry and remote sensing (RS), such as satellite and aerial-based images for urban buildings\cite{zhang2022building}.

\begin{figure}[t!]
	\centering
	\includegraphics[width=0.45\textwidth]{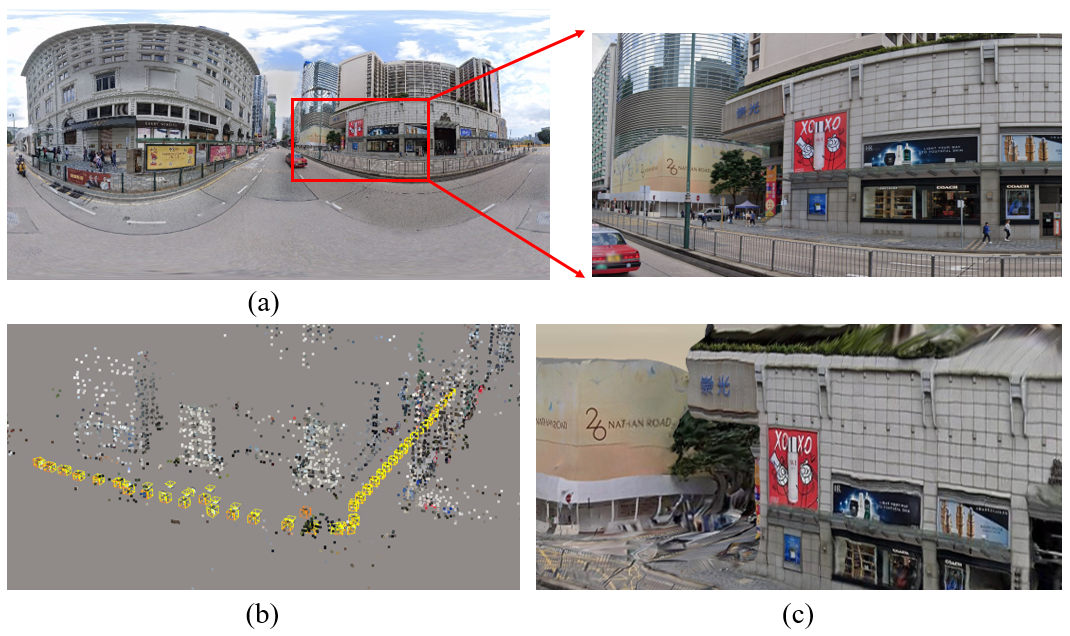}
	\caption{3D reconstruction based on spherical images. (a) a spherical image and one enlarged region for detail comparison; (b) the sparse reconstruction model; and (c) the dense reconstruction model after texture mapping.}
	\label{fig:figure1}
\end{figure}

Due to the relatively high acquisition altitudes and nearly vertical viewpoints, satellite and aerial-based RS images mainly observe building roofs over large-scale regions. With the increasing demands for fine-scale modeling, such as building facades and indoor environments, recent years have witnessed explosive development of 3D reconstruction based on low-altitude unmanned aerial vehicles (UAV)\cite{jiang2021unmanned,li2022optimized} or terrestrial mobile mapping systems (MMS)\cite{puente2013review}. Compared with satellite and airborne-based RS platforms, these near-ground platforms have the advantages of flexible instrument integration and multi-view imaging, which can record the contents that cannot be observed from high altitudes. Therefore, UAV and MMS have been used as essential RS platforms for data acquisitions in urban and indoor scenes\cite{anguelov2010google}.

Perspective cameras are the most widely used sensors for image-based 3D reconstruction. However, due to the characteristics of data acquisition in street-view and indoor environments, two main issues occur for perspective cameras with their limited FOV (Field of View) \cite{da2019dense}. On the one hand, surrounding environments are different from aerial photogrammetry. In street-view and indoor environments, the trajectory of data acquisitions is limited by street structures and indoor layouts, which causes sudden viewpoint changes at turning points and track failure between camera frames\cite{ji2020panoramic}. On the other hand, the observation regions are extended from the single-direction records in aerial photogrammetry to the omnidirectional acquisitions in street-view and indoor environments. It requires more images at each camera exposure position and increases acquisition time consumption\cite{zhang2016benefit}. Thus, effective imaging techniques are needed for 3D reconstruction in street-view and indoor situations.

Spherical cameras, also termed 360 cameras or omnidirectional cameras, can record all surrounding environments using one camera exposure. In contrast to traditional perspective cameras, recorded images of spherical cameras can cover the whole scene, whose FOV ranges are 360 and 180 degrees in horizontal and vertical directions, respectively. Due to the advantage of spherical cameras, spherical images have been adopted for 3D modeling in street-view and indoor environments\cite{bruno2019accuracy,fangi2018improving}. In addition, low-cost consumer-grade spherical cameras like the Insta360 and Ricoh theta\cite{gao2022review} are growing in popularity, which greatly simplifies data acquisition and encourages their use in a variety of fields, such as damaged building inspection\cite{jhan2021integrating}, urban environment analysis\cite{biljecki2021street}, urban geo-localization\cite{cheng2018crowd,wen2020urbanloco}, and heritage modeling\cite{fangi2013photogrammetric}. Thus, spherical images have become one of the important data sources for 3D reconstruction, especially for street-view and indoor environments \cite{kang2020review}, as illustrated in Fig. \ref{fig:figure1}.

Spherical images, however, have different characteristics when compared with traditional perspective images in the context of image-based 3D reconstruction\cite{pagani2011structure}. One of the most important differences is the camera imaging model. Consequently, 3D reconstruction from spherical images has technique differences from perspective images. In addition, fewer commercial and open-source solutions are designed for the 3D reconstruction from spherical images when compared with perspective images. Therefore, this study aims to give a review of reported techniques related to 3D reconstruction from spherical images. The main contributions of this study include: (1) we give a systematic and extensive review of recent techniques for 3D reconstruction from spherical images; (2) we present the most promising applications related to 3D reconstruction from spherical images; and (3) we also conclude the prospects for 3D reconstruction from spherical images from the aspects of technique development and application promotion. The purpose of this study is to provide useful clues to guide further research for 3D reconstruction from spherical images.

This paper is organized as follows. The state-of-the-art of data acquisition, image matching, image orientation, and dense matching for 3D modeling of spherical images is reviewed in Section \ref{sec:2}. Section \ref{sec:3} examines prospective applications for 3D reconstruction from spherical images, which is followed by the primary prospects presented in Section \ref{sec:4}. Finally, Section \ref{sec:5} concludes this work and future studies.

\section{Techniques}
\label{sec:2}

This section presents the recent techniques for 3D reconstruction from spherical images. The main workflow for the photogrammetric 3D reconstruction is first introduced, followed by data acquisition with varying spherical cameras, image matching for establishing correspondences, image orientation for estimating camera poses, and dense matching for producing point clouds. The details are listed as follows.

\subsection{Main workflow}
\label{sec:2.1}

\begin{figure}[!t]
	\centering
	\includegraphics[width=0.48\textwidth]{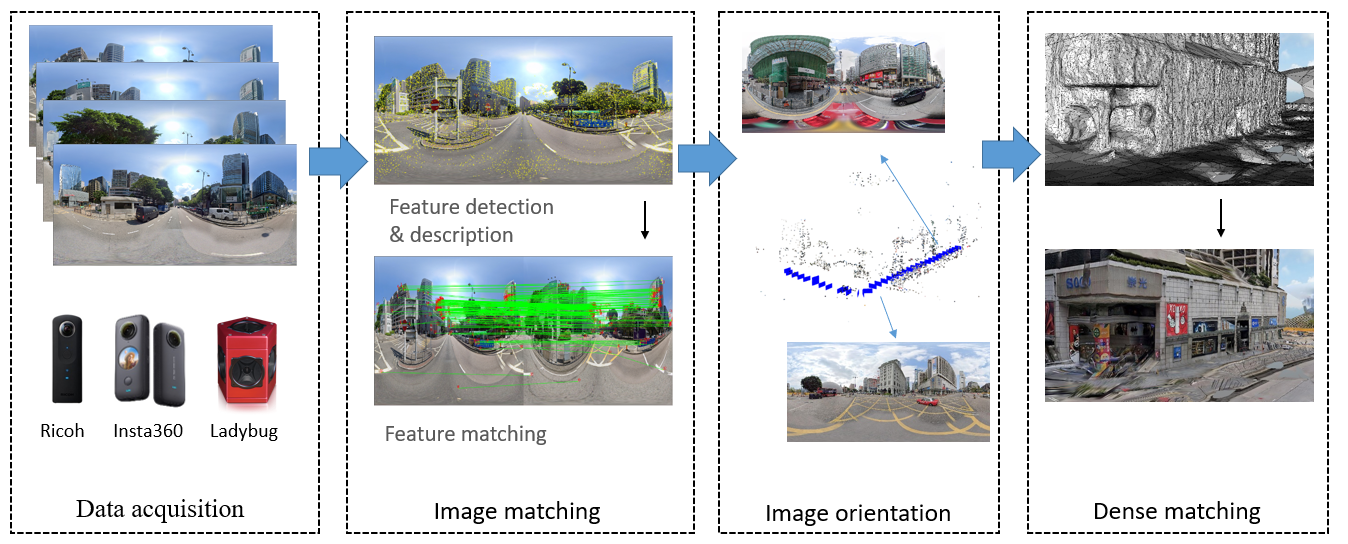}
	\caption{ The main workflow of 3D reconstruction for spherical images.}
	\label{fig:figure2}
\end{figure}

According to the literature, there are five significant steps in the workflow of image-based 3D reconstruction, i.e., data acquisition using photogrammetric systems, image matching to establish correspondences, image orientation to determine camera poses, dense matching to produce point clouds, and point cloud meshing and texturing. Spherical images differ in the camera imaging model and image representation format compared with classical perspective images, which causes extra considerations in the first four steps. Thus, this study reviews data acquisition, image matching, image orientation, and dense matching, and the main workflow of 3D reconstruction for spherical images is presented in Fig. \ref{fig:figure2}.

\subsection{Data acquisition}
\label{sec:2.2}

Data acquisition is the first step in the main workflow of 3D reconstruction for spherical images. In this section, three topics related to data acquisition are presented, i.e., spherical cameras, image representations, and acquisition platforms. The details are shown in the following subsections.

\subsubsection{ Spherical camera}
\label{sec:2.2.1}

The development of spherical cameras can be traced back to two centuries ago\cite{luhmann2004historical}, which are invented for the documentation of ancient buildings and cultural heritages. For the purpose of surveying and mapping, spherical cameras were first used in close-range photogrammetry, in which spherical images were usually captured by rotating camera around the projection center or stitching overlapped images through image matching. For aerial photogrammetry, spherical cameras are designed as an integrated instrument that consists of several well-calibrated perspective cameras.

\begin{figure}[!b]
	\centering
	\includegraphics[width=0.45\textwidth]{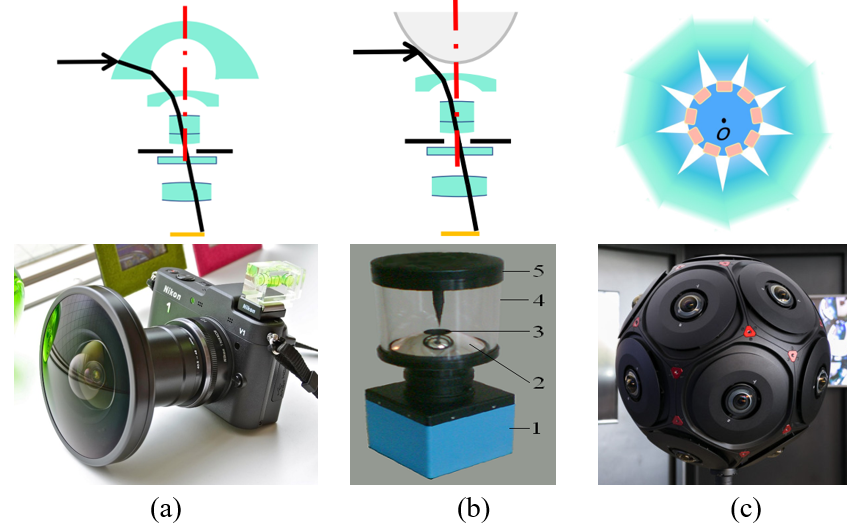}
	\caption{The design principle of three commonly used spherical cameras. (a) dioptric camera; (b) catadioptric camera; and (c) polydioptric camera.}
	\label{fig:figure3}
\end{figure}

\begin{table*}[!h]
	\caption{The detailed configurations of well-known spherical cameras. Noticeably, the resolution for Ladybug5+ is the image dimension of an individual camera.}
	\label{tab:table1}
	\centering
	\makebox[\linewidth]{
		\begin{tabular}{l l l l l l l}
			\toprule
			\textbf{Camera} &
			\textbf{Manufacturer} &
			\textbf{Principle} &
			\textbf{Number of lenses} &
			\textbf{Resolution(pixel)} &
			\textbf{Weight (kg)} &
			\textbf{Professional}\\
			\midrule
			Gear360	    & Samsung          & polydioptric  & 2	 & 4096×2048   & 0.13  & $\times$  \\
			Theta Z1    & Ricoh            & polydioptric  & 2   & 6720×3360   & 0.18  & $\times$  \\
			Theta X	    & Ricoh            & polydioptric  & 2   & 11008×5504  & 0.17  & $\times$  \\
			Max 360	    & GoPro            & polydioptric  & 3   & 5760×2880   & 0.16  & $\times$  \\
			Sphere      & Insta360         & polydioptric  & 2   & 6080×3040   & 0.19  & $\times$  \\
			Pro 2       & Insta360         & polydioptric  & 6   & 7680×7680   & 1.55  & $\checkmark$  \\
			Panono      & Professional360  & polydioptric  & 36  & 16000×8000  & 0.48  & $\checkmark$  \\
			Ladybug5+   & Teledyne FLIR    & polydioptric  & 6   & 2048×2464*  & 3.00  & $\checkmark$  \\
			Civetta     & Weiss AG         & dioptric      & /	 & 230 M	   & 5.70  & $\checkmark$  \\
			\bottomrule
	\end{tabular}}
\end{table*}

In recent years, the performance of spherical cameras has been greatly improved by the progress in the fields of camera sensors and image processing techniques, e.g., high-resolution digital cameras and high-precision stitching algorithms. Based on the design principle, spherical cameras can be divided into three major categories, i.e., dioptric cameras, catadioptric cameras, and polydioptric cameras\cite{gao2022review,scaramuzza2014omnidirectional}:

\begin{itemize}
	\item \textbf{Dioptric cameras} use a particular lens group to refract rays that compress the direction of the light entering the subsequent lens group. The obtained FOV reaches 360 degrees in the horizontal direction and is larger than 90 degrees in the vertical direction. Thus, two lenses combined back to back can capture the full surroundings. Fig. \ref{fig:figure3}(a) is an example of the fisheye camera.
	\item \textbf{Catadioptric cameras} utilize the combination of a lens group for ray refraction and a special mirror for ray reflection, e.g., a parabolic, hyperbolic, or elliptical mirror, with a standard camera to achieve the FOV of 360 degrees and greater than 100 degrees in the horizontal and vertical direction, respectively. Compared with dioptric cameras, this design can reflect the surrounding light into the subsequent lens. Fig. \ref{fig:figure3}(b) is an illustration of the catadioptric camera.
	\item \textbf{Polydioptric cameras} adopt multiple dioptric cameras to obtain a real spherical FOV, i.e., 360 degrees and 180 degrees in the horizontal and vertical directions, respectively, in which dioptric cameras have overlapping FOV to facilitate image stitching. Fig. \ref{fig:figure3}(c) shows an example of a polydioptric camera comprising 16 cameras.
\end{itemize}

\begin{figure}[!h]
	\centering
	\subfloat[]{\includegraphics[height=0.15\textwidth]{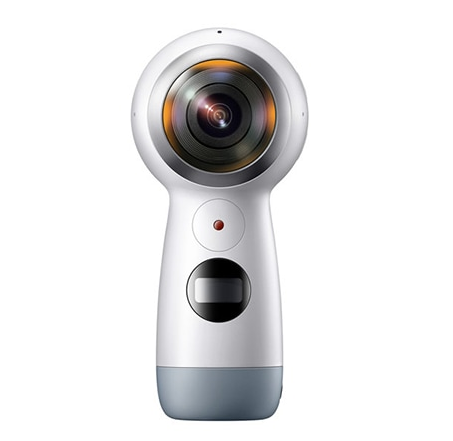}%
		\label{fig:figure4-a}}
	\subfloat[]{\includegraphics[height=0.15\textwidth]{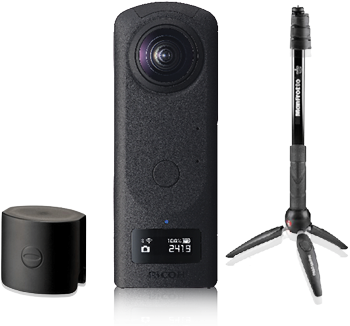}%
		\label{fig:figure4-b}}
	\subfloat[]{\includegraphics[height=0.15\textwidth]{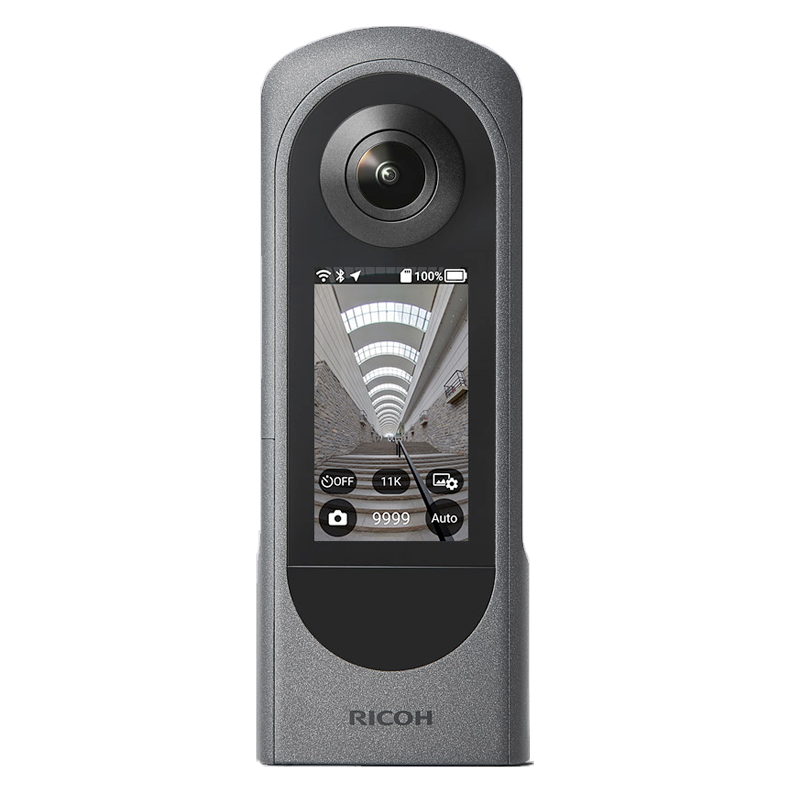}%
		\label{fig:figure4-c}}	\\
	\subfloat[]{\includegraphics[height=0.15\textwidth]{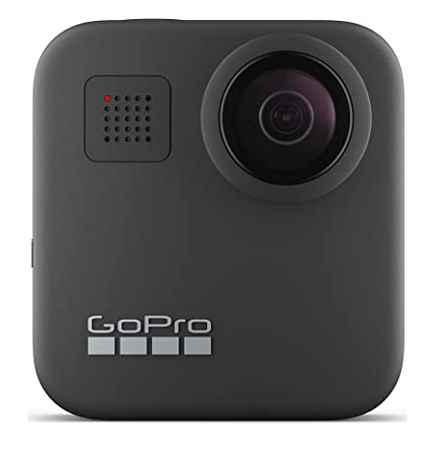}%
		\label{fig:figure4-d}}
	\subfloat[]{\includegraphics[height=0.15\textwidth]{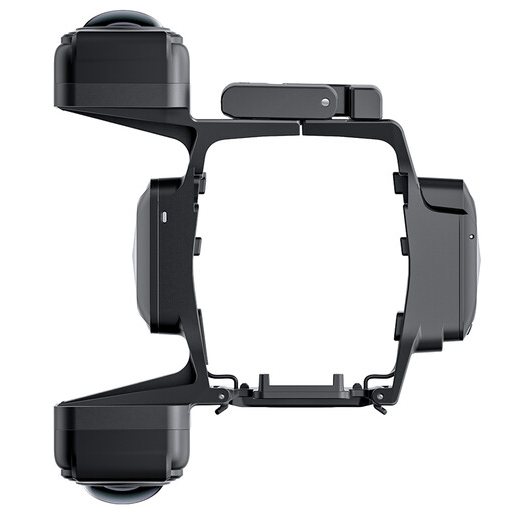}%
		\label{fig:figure4-e}}
	\subfloat[]{\includegraphics[height=0.15\textwidth]{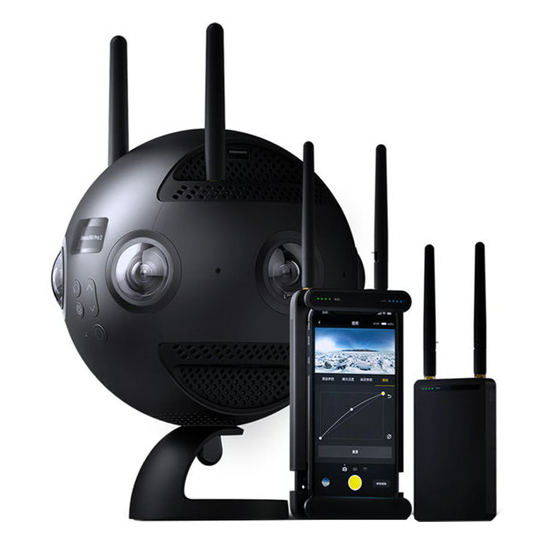}%
		\label{fig:figure4-f}}  \\
	\subfloat[]{\includegraphics[height=0.15\textwidth]{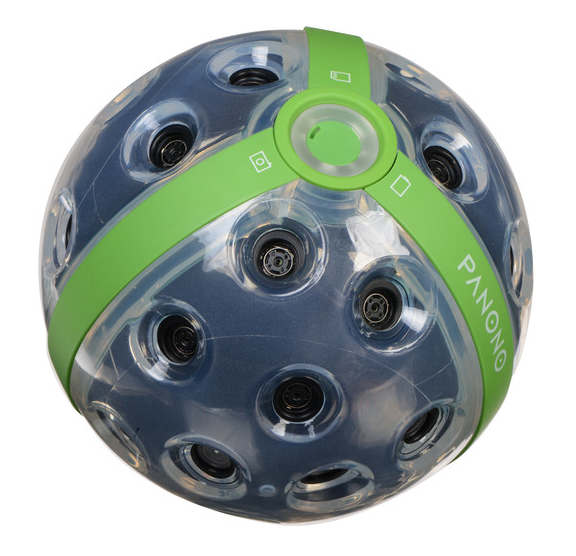}%
		\label{fig:figure4-g}}
	\subfloat[]{\includegraphics[height=0.15\textwidth]{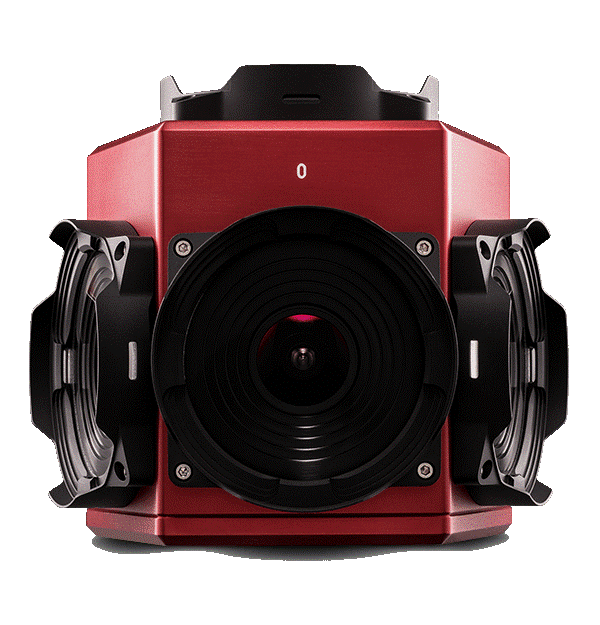}%
		\label{fig:figure4-h}}
	\subfloat[]{\includegraphics[height=0.15\textwidth]{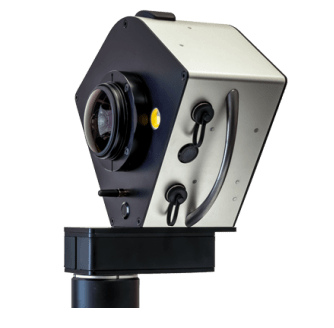}%
		\label{fig:figure4-i}}
	\caption{The illustration of the well-known spherical cameras. (a) Gear360; (b) Theta Z1; (c) Theta X; (d) Max 360; (e) Sphere; (f) Pro 2; (g) Panono; (h) Ladybug5+; (i) Civetta.}
	\label{fig:figure4}
\end{figure}

Among the three categories, polydioptric cameras have become the most extensively used spherical cameras for both professional and consumer-grade applications because of two main reasons. On the one hand, it can provide the full omnidirectional imaging technique; on the other hand, recorded images have extremely high resolutions due to the use of multiple cameras. Table \ref{tab:table1} presents the detailed configurations of well-known spherical cameras, including the consumer-grade cameras, e.g., the Ricoh Theta series and Insta360 Sphere, and the professional cameras, e.g., Insta360 Pro2 and Teledyne FLIR Ladybug5+. Except for Weiss AG Civetta, all the other spherical cameras are designed by using the polydioptric mechanism. For consumer-grade cameras, the lens number is usually configured as 2 or 3. The illustration of the cameras is presented in Fig. \ref{fig:figure4}. In general, there are two ways to increase the resolution of recorded images, i.e., increasing the number of integrated cameras or exchanging the style of image recording. The former has been used in Panono and Ladybug5+, which enables instant image recording, e.g., equipped with an MMS system. The latter has been used in Weiss AG Civetta to obtain extremely high resolution. This strategy, however, sacrifices the capability of instant acquisition, and it is more suitable for site-based image recording.

\subsubsection{Image representation}
\label{sec:2.2.2}

\begin{figure}[t!]
	\centering
	\includegraphics[width=0.48\textwidth]{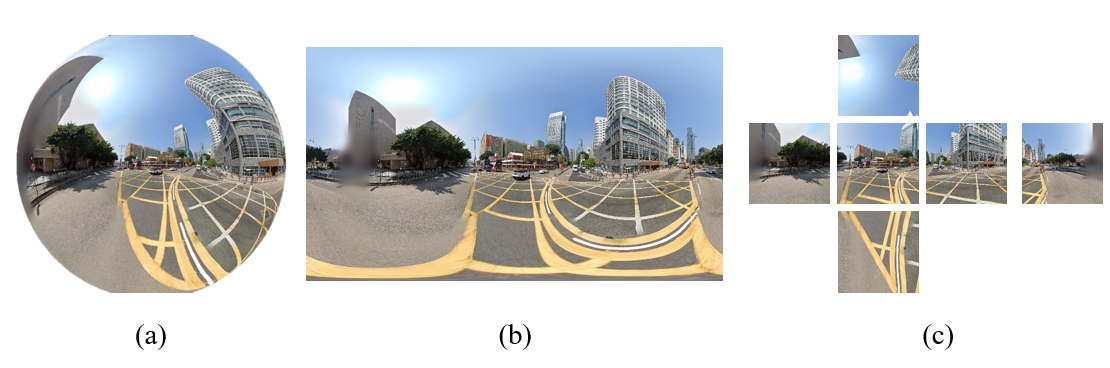}
	\caption{Three typical representation for spherical images: (a) spherical representation; (b) equirectangular representation; (c) cubic-map representation.}
	\label{fig:figure5}
\end{figure}

Spherical images record the surrounding environments at each camera exposure position. In contrast to the 2D plane representation of perspective images, Fig. \ref{fig:figure5} presents the most widely used three types of image representation methods\cite{da20223d}.

\begin{itemize}
	\item The first one is the spherical representation. Objects in surrounding environments are mapped onto a sphere, as presented in Fig. \ref{fig:figure5} (a). Spherical representation is useful for panoramic navigation, which has been widely used for street-view navigation, e.g., Google and Baidu Street View. Spherical representation, however, is unsuitable for image processing and hardware storage.
	\item Equirectangular images, which are produced by the equirectangular projection (ERP), are a common representation, as shown in Fig. \ref{fig:figure5} (b). Similar to perspective images, equirectangular images can be considered typical images and processed by existing algorithms, e.g., feature extraction and matching\cite{pagani2011structure}. Because of the projection from 3D sphere to 2D plane, geometric distortions are introduced to equirectangular images, especially for the regions near sphere poles.
	\item The third image representation, i.e., cubic-map representation (CMP), is created to alleviate the distortion in equirectangular projection, which converts one spherical image into six concentric perspective images, as presented in Fig. \ref{fig:figure5} (c). Each cubic-map image can be considered as one typical perspective image whose projection distortions have been removed. However, this representation would decrease the overlap region between frames and increase the image number for image orientation.
\end{itemize}

Since the simple and typical format, equirectangular representation has been extensively adopted for spherical images, including well-known open-source and commercial software packages, e.g., OpenMVG\cite{moulon2016openmvg}, Agisoft Metashape \cite{2022Agisoft}, and Pix4dMapper \cite{2022Pix4dMapper}. Thus, this review pays more attention to the equirectangular representation of spherical images.

\subsubsection{Acquisition platform}
\label{sec:2.2.3}

\begin{figure}[!t]
	\centering
	\includegraphics[width=0.45\textwidth]{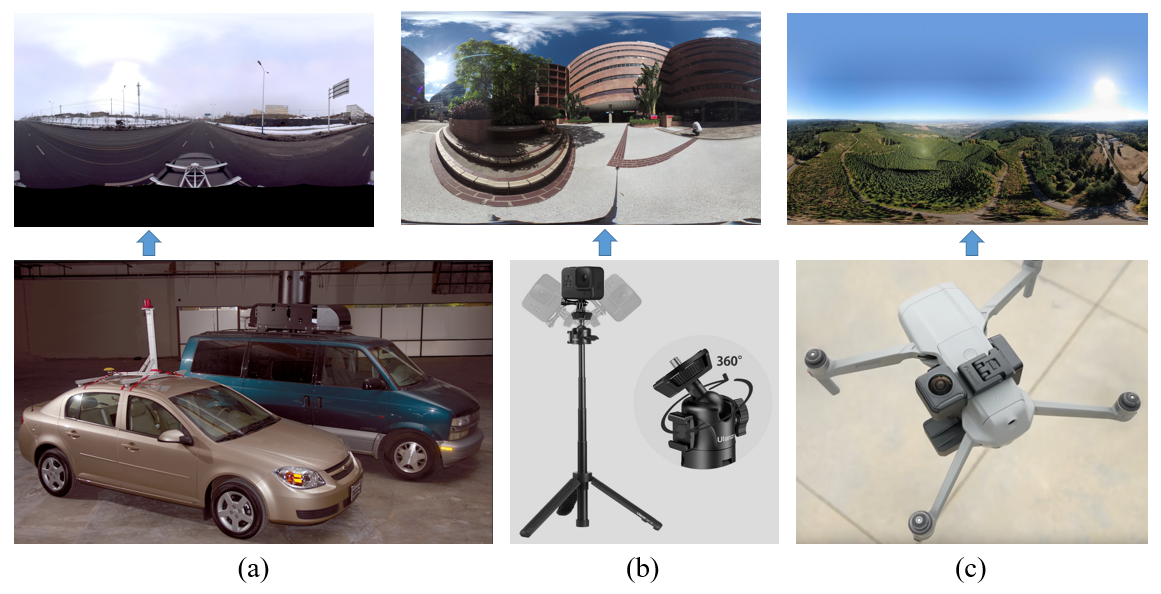}
	\caption{Different acquisition platforms and corresponding sample images. (a) moving vehicles; (b) ground-fixed tripods; and (c) unmanned aerial vehicles.}
	\label{fig:figure6}
\end{figure}

Depending on their individual properties, spherical cameras can be operated manually or with various remote sensing platforms. The most extensively adopted platforms can be moving vehicles\cite{anguelov2010google}, ground-fixed tripods\cite{herban2022use}, and handheld poles. These platforms record images along urban streets or around center landmarks, as presented in Fig. \ref{fig:figure6}(a) and Fig. \ref{fig:figure6}(b).

\begin{figure}[!t]
	\centering
	\includegraphics[width=0.45\textwidth]{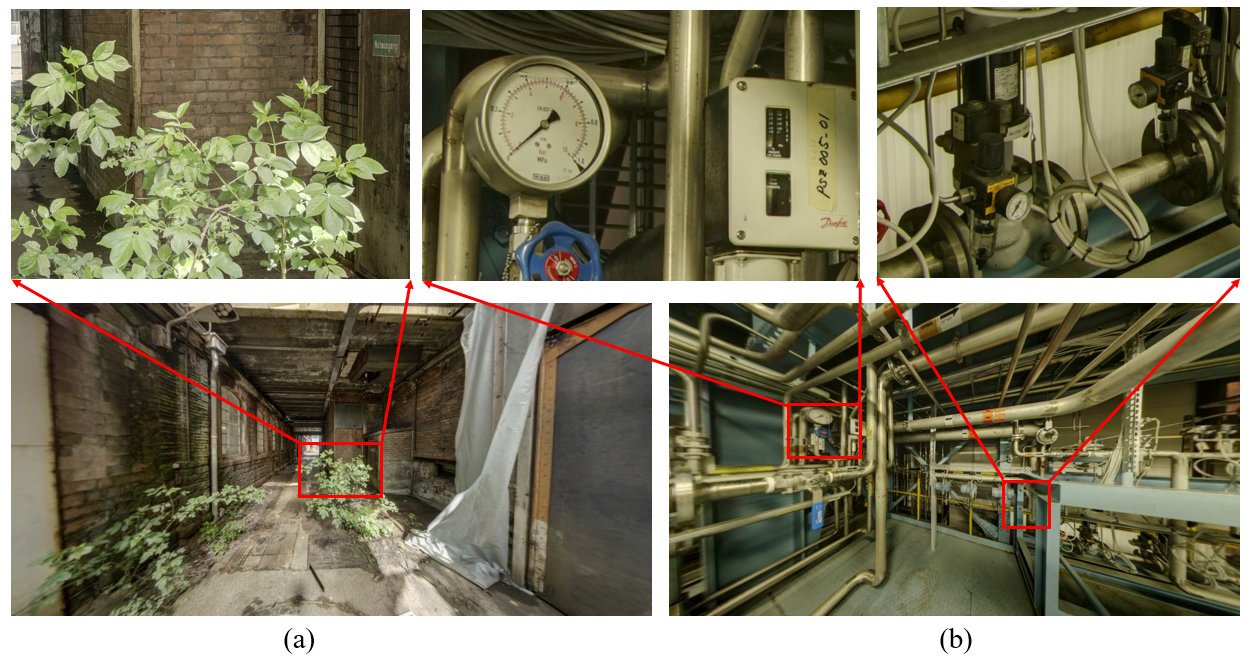}
	\caption{Spherical images recorded by the Weiss AG Civetta camera \cite{2022WeissAg}. (a) the example for the outdoor scene; and (b) the example for the indoor scene.}
	\label{fig:figure7}
\end{figure}

In recent years, UAV platforms are also designed to accommodate spherical cameras\cite{zhang2020uav}, which can record spherical images from relatively high altitudes, as shown in Fig. \ref{fig:figure6}(c). Compared with spherical cameras for the other two platforms, e.g., Ladybug5+ for moving vehicles and Civetta for ground-fixed tripods, spherical cameras for UAVs are strict to the weight and dimension of sensors due to the limited payload weight and flight endurance. In addition, spherical cameras are required instant recording ability when mounted on moving vehicles and UAV platforms, which would restrict the resolution of recorded images. On the contrary, spherical cameras designed for ground-fixed tripods can record images with extremely high spatial resolution. Fig. \ref{fig:figure7} shows the images recorded by the Weiss AG Civetta spherical camera, from which details can be observed from both outdoor and indoor recorded spherical images. In the literature, there exist some useful and public datasets captured by sphere cameras, which is presented in Table \ref{tab:table8}.

\begin{table*}[t!]
	\caption{A list of benchmark datasets for spherical images.}
	\label{tab:table8}
	\centering
	\makebox[\textwidth]{
		\begin{tabular}{p{0.15\textwidth} p{0.05\textwidth} p{0.2\textwidth} p{0.4\textwidth}}
			\toprule
			\textbf{Name}		& 
			\textbf{Year}       &
			\textbf{Purpose}    &
			\textbf{Website}		\\
			\midrule
			\multirow{2}{0.15\textwidth}{3D60}
			& 2021
			& Dense matching
			& \url{https://vcl3d.github.io/Pano3D}\\
			& \multicolumn{3}{L{0.6\textwidth}}{It provides composited and realistically scanned 3D datasets of interior spaces that are used to generate high-quality, densely annotated spherical panoramas\cite{zioulis2018omnidepth}.} \\
			\midrule
			\multirow{2}{0.15\textwidth}{Re-rendering subsets of Stanford 2D-3D and Matterport}
			& 2020
			& Dense matching
			& \url{https://albert100121.github.io/360SD-Net-Project-Page/}\\
			& \multicolumn{3}{L{0.6\textwidth}}{It contains re-rendered subsets of the Stanford 2D-3D and Matter-port3D data sets, which consists of 3,577 1024 x 512 stereoscopic ERP images\cite{wang2020360sd}.} \\
			
			\midrule
			\multirow{2}{0.15\textwidth}{Structured3D}
			& 2020
			& Structured 3D modeling
			& \url{https://structured3d-dataset.org/}\\
			& \multicolumn{3}{L{0.6\textwidth}}{It provides 196,515 frames from 3,500 synthetic, photo-realistic house designs, and each scene has 1024 × 512 ERP images at three different light conditions and camera poses\cite{chang2017matterport3d}.} \\
			
			\midrule
			\multirow{2}{0.15\textwidth}{Re-rendering subset of Stanford 2D-3D}
			& 2019
			& Dense matching
			& \url{https://github.com/pokonglai/ods-net}\\
			& \multicolumn{3}{L{0.6\textwidth}}{It provides a re-rendered subset of the Stanford 2D-3D dataset. It provides about 50,000 256 × 128 stereo ERP images separated by a 6.5cm horizontal baseline\cite{lai2019real}.} \\
			
			\midrule
			\multirow{2}{0.15\textwidth}{PanoSUNCG}
			& 2018
			& Camera motion and depth
			& \url{https://fuenwang.ml/project/360-depth/} \\
			& \multicolumn{3}{L{0.6\textwidth}}{It provides about 25,000 images captured by SUNCG from 103 different scenes in five camera paths, including color, depth, and rendering tracks\cite{wang2018self}.}\\
			
			\midrule
			\multirow{2}{0.15\textwidth}{Stanford 2D-3D}
			& 2017
			& Dense matching
			& \url{http://3dsemantics.stanford.edu} \\
			& \multicolumn{3}{L{0.6\textwidth}}{It contains 1,413 full Field of View indoor realistic capture of six wide range of areas, which provides depth, normal, and semantic mapping\cite{armeni2017joint}.}\\
			
			\midrule
			\multirow{2}{0.15\textwidth}{Matterport3D}
			& 2017
			& Dense matching
			& \url{https://niessner.github.io/Matterport/} \\
			& \multicolumn{3}{L{0.6\textwidth}}{It provides 10,800 CMP panoramas from 90 real building scale scenes, including depth, camera position, and semantic segmentation\cite{chang2017matterport3d}.}\\
			\bottomrule
		\end{tabular}
	}
\end{table*}

\subsection{Image matching}
\label{sec:2.3}

Image matching is the second step in the main workflow of 3D reconstruction for spherical images, which aims to establish correspondence matches between image pairs with high inlier ratio and even spatial distribution. In the literature, image matching has been a well-studied topic in the fields of photogrammetry and computer vision, which can be verified from earlier hand-crafted algorithms to recent learning-based networks\cite{hartmann2016recent,jiang2021learned,ma2021image}. In this section, we focus on reported methods that can be utilized for spherical images. Table \ref{tab:table2} lists the algorithms for image matching.

\begin{table*}[t!]
	\caption{The algorithms for feature matching of spherical images.}
	\label{tab:table2}
	\centering
	\makebox[\textwidth]{
		\begin{tabular}{p{0.15\textwidth} p{0.1\textwidth} p{0.1\textwidth} p{0.4\textwidth}}
			\toprule
			\textbf{Name}			& 
			\textbf{Language}       &
			\textbf{Year}           &
			\textbf{Website}		\\
			\midrule
			\multirow{2}{0.1\textwidth}{OmniCV}
			& C++
			& 2020
			& \url{https://github.com/kaustubh-sadekar/OmniCV-Lib}\\
			& \multicolumn{3}{L{0.6\textwidth}}{A computer vision library for omnidirectional cameras, e.g., dioptric, catadioptric and polydioptric cameras. It provides tools for format conversion and image viewing.} \\
			
			\midrule
			\multirow{2}{0.1\textwidth}{SPHORB}
			& C++
			& 2015
			& \url{https://github.com/tdsuper/SPHORB}\\
			& \multicolumn{3}{L{0.6\textwidth}}{A package based on a nearly regular hexagonal grid parametrization of the sphere geodesic grid, which adapts planar ORB to the spherical domain\cite{zhao2015sphorb}.} \\
			
			\midrule
			\multirow{2}{0.1\textwidth}{SSIFT}
			& Matlab
			& 2009
			& \url{https://github.com/Artcs1/Keypoints}\\
			& \multicolumn{3}{L{0.6\textwidth}}{It transforms the plane SIFT to spherical coordinates and proposes two descriptors for feature matching\cite{cruz2012scale}. Please refer to the file SSIFT.m in this repository.} \\
			
			\midrule
			\multirow{2}{0.1\textwidth}{Tangent Images}
			& Python
			& 2020
			& \url{https://github.com/meder411/Tangent-Images}\\
			& \multicolumn{3}{L{0.6\textwidth}}{Distortion is alleviated by rendering the spherical image as a set of local planar image grids tangent to the subdivided icosahedron\cite{eder2020tangent}.} \\
			
			\midrule
			\multirow{2}{0.1\textwidth}{UGSCNN}
			& Python
			& 2019
			& \url{https://github.com/maxjiang93/ugscnn}\\
			& \multicolumn{3}{L{0.6\textwidth}}{A CNN network for spherical signals based on parameterized differential operators on unstructured grids\cite{jiang2019spherical}.}\\
			
			\midrule
			\multirow{2}{0.1\textwidth}{DEEPSPHERE}
			& Python
			& 2020
			& \url{https://github.com/deepsphere}\\
			& \multicolumn{3}{L{0.6\textwidth}}{A graphical representation of a sampling sphere that achieves a controlled balance between efficiency and rotationally equivalent variance\cite{defferrard2020deepsphere}.} \\
			
			\midrule
			\multirow{2}{0.1\textwidth}{S2CNN}
			& Python
			& 2018
			& \url{https://github.com/jonkhler/s2cnn}\\
			& \multicolumn{3}{L{0.6\textwidth}}{A library for the rotation equivariant CNNs for spherical signals (e.g. omnidirectional images, signals on the globe)\cite{cohen2018spherical}.} \\
			
			\midrule
			\multirow{2}{0.1\textwidth}{SphereNet}
			& Python
			& 2018
			& \url{https://github.com/ChiWeiHsiao/SphereNet-pytorch}\\
			& \multicolumn{3}{L{0.6\textwidth}}{A network that adjusts sampling positions of CNN kernels, which can transfer existing perspective networks to the omnidirectional case\cite{coors2018spherenet}.} \\
			\bottomrule
		\end{tabular}
	}
\end{table*}

\subsubsection{Feature detection and description}
\label{sec:2.3.1}

The purpose of feature detection and description is to detect distinguishable keypoints that can be found in overlapped images and calculate their robust descriptors that are invariant to the changes in scale, viewpoint, and illumination. Due to the equirectangular projection of spherical images, more serious geometric distortions disturb feature detection and description. According to the strategy used, existing methods can be divided into four groups, including classical methods, 2D plane-based methods, 3D sphere-based methods, and learning-based methods.

(1) Classical methods

Classical feature detection and description methods can also be applied to equirectangular images because of two main reasons. On the one hand, image regions near the sphere equator have relatively small distortions after equirectangular projection; on the other hand, most data acquisition campaigns are conducted by fixing the roll and pitch angles of sphere cameras, such as cameras mounted on moving vehicles for street-view images\cite{torii2009google} or on fixed tripods for corridor photos\cite{herban2022use}. These two conditions ensure the repeatability of extracted features on equirectangular images.

\begin{figure}[!t]
	\centering
	\includegraphics[width=0.45\textwidth]{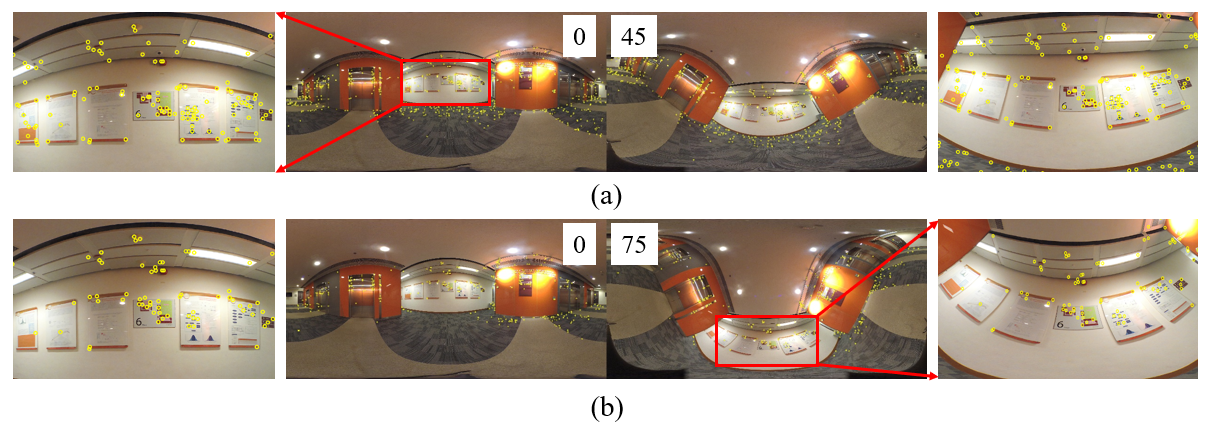}
	\caption{The influence of geometric distortion on feature matching using the classical SIFT. The value in each image indicates the rotation angle around the X axis, and the yellow circles represent the matched feature points between the two images.}
	\label{fig:figure8}
\end{figure}

In the literature, classical feature detectors have been applied to equirectangular images, including floating and binary feature descriptors\cite{bay2008speeded,lowe2004distinctive,morel2009asift,rublee2011orb}. \cite{torii2009google} used SURF (Speeded-up Robust Features) to extract features from street-view images and conduct image orientation based on SfM (Structure from Motion). Thus, existing algorithms can be directly utilized for equirectangular images, which can achieve high efficiency without extra computational costs. Fig. \ref{fig:figure8} gives an example of feature extraction using the classical SIFT.

%For the image pair shown in Fig. \ref{fig:figure8}(a) and Fig. \ref{fig:figure8}(b), a total number of 746 and 392 correspondences are matched respectively.

(2) 2D plane-based methods

When spherical images are rotated around the X (pitch angle) or Z (roll angle) axis, the image appearances near the sphere equator would change obviously, which causes serious geometric distortions. For visual analysis, Fig. \ref{fig:figure8} illustrates three images that are captured by rotating around the X axis with the angle of 0, 45, and 75 degrees, respectively. It is clearly shown that geometric distortions increase dramatically with the increase of rotation angles. Because of the introduced distortions, the number of correspondences decreases in these two corresponding regions. Feature detection and description must therefore receive more attention.
In the field of photogrammetry, image rectification has been widely used to decrease the influence of geometric distortions on feature detection and description \cite{jiang2017board}. Similar to the rectification strategy, some research has been documented for spherical images, which can be divided into three main groups, i.e., global methods, semi-global methods, and local methods.

\begin{figure}[!t]
	\centering
	\includegraphics[width=0.45\textwidth]{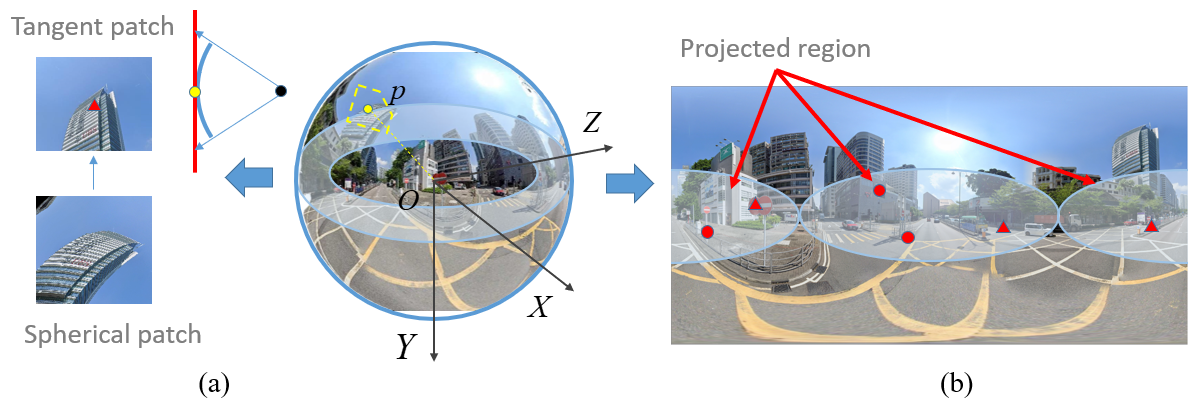}
	\caption{The illustration of 2D plane-based methods. (a) local methods based on the tangent plane projection of feature point patches; (b) semi-global methods based on the projection around the sphere equator region. Red shapes indicate detected features.}
	\label{fig:figure9}
\end{figure}

\begin{itemize}
	\item \textbf{Global methods} aim to rectify the whole image, and cubic-map representation has been the classical strategy, as presented in Fig. \ref{fig:figure5}(c). For example, \cite{wang2018cubemapslam} proposed reprojecting spherical images into cubic-map images for feature detection and implementing a SLAM (Simultaneous Localization and Mapping) system, termed CubemapSLAM.
	
	\item \textbf{Semi-global methods} are designed to reproject a proportion of spherical images and conduct feature detection on the rectified image region, which is inspired by the truth that small distortions exist near the sphere equator, as presented in Fig. \ref{fig:figure9}(b). \cite{taira2015robust} proposed generating three reprojected equirectangular images by a rotation of 0, 60, and 120 degrees around X axis, respectively, and detecting features from the equator regions in each rectified equirectangular image.
	
	\item \textbf{Local methods} aim to rectify local image regions around detected feature points and calculate their descriptors on the rectified image patches\cite{eder2020tangent,chuang2018rectified}, as presented in Fig. \ref{fig:figure9}(a). In the work of \cite{chuang2018rectified}, image patches around feature points are projected to the tangent planes across these feature points, and SURF descriptors are computed from these patches with similar perspective views. Although 2D plane-based methods are based on a simple principle, extra time costs are required in image rectification.
\end{itemize}

(3) 3D sphere-based methods

\begin{figure}[!t]
	\centering
	\includegraphics[width=0.45\textwidth]{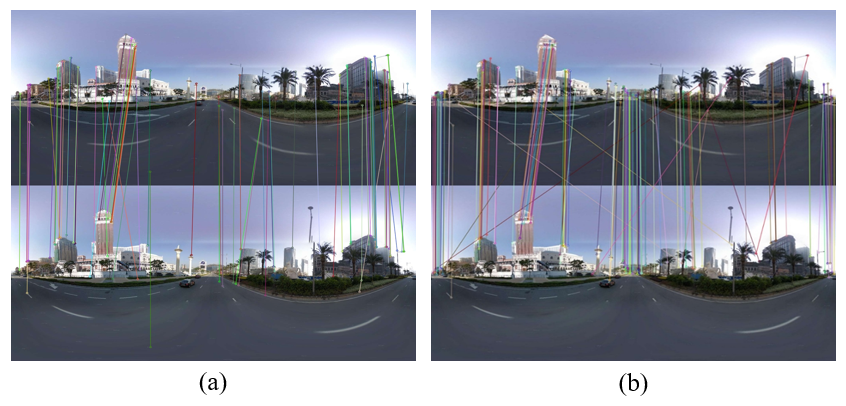}
	\caption{The comparison of feature detection and matching between ORB and SPHORB. (a) the matching result of ORB; (b) the matching result of SPHORB.}
	\label{fig:figure10}
\end{figure}

In contrast to image geometric rectification, some research starts from scratch to design and implement algorithms for feature detection and description of spherical images\cite{cruz2012scale,guan2017brisks,zhao2015sphorb}. The primary motivation is to avoid the high computational costs consumed in image geometric rectification and decrease the influence of geometric distortions introduced by equirectangular projection. Naturally, the spherical representation is the optimal solution instead of the equirectangular representation used in the above-mentioned methods. Thus, existing research exploits the spherical coordinate to design new feature detection and description algorithms. The core of these methods is how to construct the scale space pyramid on spherical images\cite{arican2012scale}. Like SIFT, \cite{cruz2012scale} implemented a spherical SIFT, termed SSIFT, which directly simulates image representation, scale space construction, extreme point detection, and descriptor calculation in the spherical coordinates, and shows better performance compared with classical SIFT. Inspired by this work, others attempt to improve the efficiency of SSIFT by replacing the time-consuming SIFT with binary detectors and descriptors, such as SPHORB\cite{zhao2015sphorb} and BRISKS\cite{guan2017brisks}. Fig. \ref{fig:figure10} shows feature matching of ORB and SPHORB. In a word, 3D sphere-based methods can avoid the geometric distortions in equirectangular images with the sacrifice of efficiency due to the computation on the sphere.

(4) Learning-based methods

\begin{figure}[!t]
	\centering
	\includegraphics[width=0.45\textwidth]{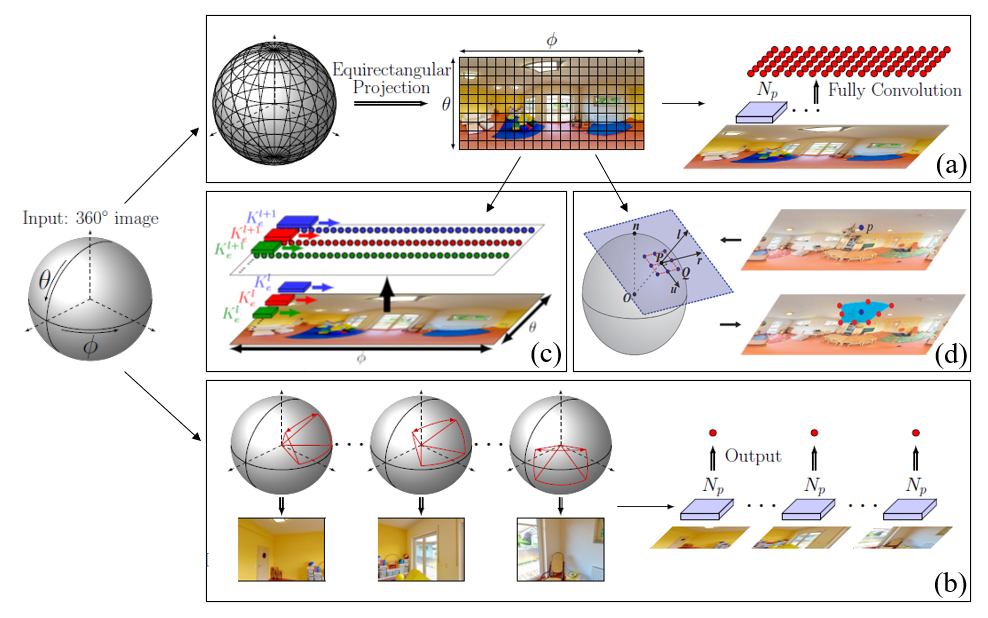}
	\caption{The principle of CNN for spherical images\cite{su2017learning},\cite{zhao2018distortion}. (a) direct applying CNN on the ERP spherical images; (b) applying CNN on the tangent images; (c) adjusting the kernel shape of CNN at varying latitudes; (d) adjusting the sampling points of CNN kernels by using tangent plane projection.}
	\label{fig:figure11}
\end{figure}

Due to the powerful representation learning ability, CNN (Convolutional Neural Network) based deep learning networks have been extensively used for feature detection and description. According to the work of \cite{jiang2021learned}, existing networks can be divided into three groups, i.e., joint feature and metric learning networks\cite{han2015matchnet,kumar2016learning,simo2015discriminative}, separate detector and descriptor learning networks\cite{luo2019contextdesc,luo2018geodesc,mishchuk2017working},\cite{tian2017l2}, and joint detector and descriptor learning networks\cite{detone2018superpoint,Dusmanu2019CVPR}. For the first group, CNN models learn a similarity function to predict image patch similarities and integrate feature representation and matching within the same network. For the second group, CNN networks only learn feature representation, and feature matching is conducted based on classical strategies, such as the L2-norm Euclidean distance-based metric between feature descriptors. For the third group, trained models learn detectors and descriptors simultaneously, which can cope with images recorded in varying conditions. Similar to hand-crafted methods, existing networks have been used for spherical images by using pre-trained or finetuned models, as shown in Fig. \ref{fig:figure11}(a). For further details, readers can refer to\cite{da2017evaluation},\cite{murrugarra2022pose}.

Due to inevitable distortions in sphere-to-plane projection, CNN networks for perspective images may obtain inaccurate results. To address this issue, reported approaches in literature can be divided into three groups, i.e., tangent projection methods, CNN kernel shape resizing methods, and CNN sampling point adjustment methods.

\begin{itemize}
	\item For the first one, similar to 2D plane-based methods, equirectangular images are first projected to undistorted tangent images\cite{eder2020tangent} or divided into quasi-uniform discrete images\cite{shan2018descriptor}, and then existing CNN networks are applied to resulting images, as shown in Fig. \ref{fig:figure11}(b). Although these methods can achieve high accurate predictions, they suffer from high computational costs due to resampling of 3D spherical images to 2D plane images.
	
	\item For the second one, CNN networks are designed to work on equirectangular images by adjusting the CNN kernel shape\cite{coors2018spherenet},\cite{su2017learning},\cite{zhao2018distortion}. In the work of \cite{su2017learning}, a network termed SPHCONV has been proposed, which aims to produce results as the output of applying perspective CNN networks to the corresponding tangent images. SPHCONV was achieved by defining convolution kernels with varying shapes for pixels in different image rows, as illustrated in Fig. \ref{fig:figure11}(c). Similarly, \cite{su2019kernel} proposed a kernel transformer network (KTN) to learn spherical kernels by taking as input the latitude angle and source kernels for perspective images.
	
	\item For the third one, sampling points of CNN kernels are adjusted based on ERP sphere image distortions instead of adjusting convolution kernel shape. For example, \cite{zhao2018distortion} and \cite{coors2018spherenet} designed distortion-aware networks that sample non-regular grid locations according to the distortions of different pixels, as shown in Fig. \ref{fig:figure11}(d). The core idea is to determine sampling locations based on the sphere projection of a regular grid on the corresponding tangent plane. Due to regular convolution kernels, these frameworks enable the transfer between CNN models for perspective and equirectangular images.
\end{itemize}

\subsubsection{Feature matching}
\label{sec:2.3.2}

The purpose of feature matching is to search correspondences from two sets of feature descriptors. Generally, feature matching is achieved by using nearest-neighbor searching based on the L2-norm Euclidean distance metric\cite{muja2009fast}. In the literature, extensive research has been reported for feature matching from the aspects of efficiency acceleration and precision improvement\cite{hartmann2016recent},\cite{jiang2021unmanned}. Due to the high dimension of local feature descriptors and the large number of spherical images, exhaustive feature matching consumes extremely high time costs. Thus, it becomes very critical to increase the efficiency of feature matching.

\begin{figure}[!t]
	\centering
	\includegraphics[width=0.45\textwidth]{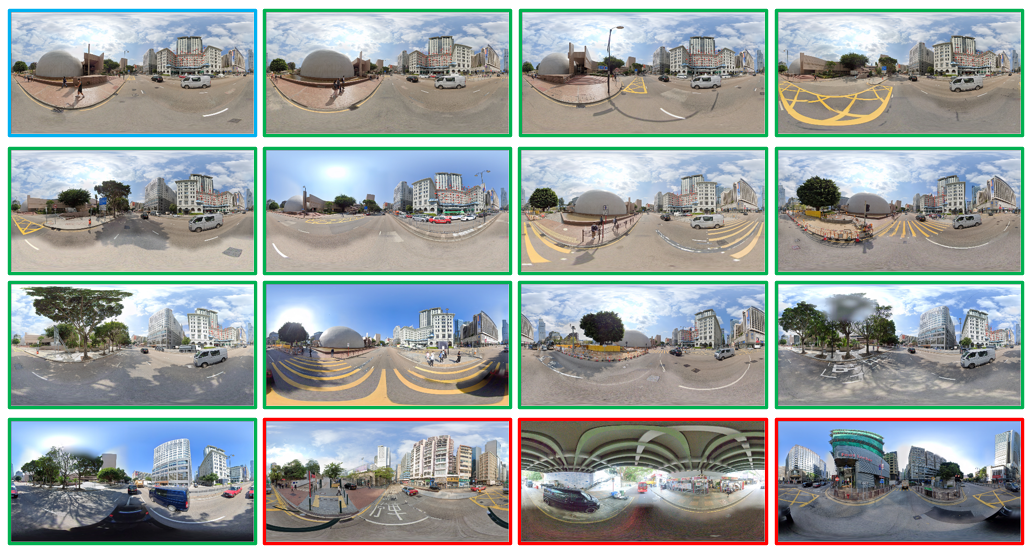}
	\caption{An example of CBIR for spherical images. The image with blue borders is the query image, and the other images are the retrieval results. Images with green and red borders represent true and false retrieval results, respectively.}
	\label{fig:figure12*}
\end{figure}

In the literature, there are valuable methods for accelerating feature matching, including restricting the number of features in feature matching, decreasing the number of images, and reducing the number of match pairs. For a detailed review, the readers can refer to this work\cite{hartmann2016recent},\cite{jiang2020effcient}. Among these reported methods, match pair selection can be the most straightforward and effective way to accelerate feature matching. Three strategies can be exploited for spherical images based on their acquisition environments. First, data acquisition constraints can be used, e.g., the time-sequential constraint for street-view images. In other words, feature matching can be restricted to neighbors according to their acquisition time. Second, for professional MMS systems and consumer-grade sensors, precise or rough GNSS (Global Navigation Satellite System) data is usually recorded simultaneously with spherical images, which provides the best clue to select spatially overlapped match pairs. Third, the CBIR (Content-based Image Retrieval) technique has become a standard module\cite{jiang2020efficient},\cite{zheng2017sift}. It merely uses images for visual retrieval without other assumptions and auxiliary data. Fig. \ref{fig:figure12*} shows an example of CBIR.

\subsection{Image orientation}
\label{sec:2.4}

Image orientation aims to estimate camera poses, scene structures, and intrinsic parameters based on established two-view correspondences. Image orientation is also termed aerial triangulation (AT) in photogrammetric 3D reconstruction, which requires good initial values of unknown parameters and well-designed configurations in data acquisition. Image orientation of spherical images can be achieved through the well-known Structure from Motion (SfM) technique for multi-view geometry in computer vision\cite{snavely2006photo} or the Simultaneous Localization and Mapping (SLAM) technique for instant localization in robot vision\cite{mouragnon2006real}. Thus, this subsection presents camera calibration, SfM-based offline, and SLAM-based online methods for spherical image orientation.

\subsubsection{Camera calibration}
\label{sec:2.4.1}

\hfill

(1) camera imaging model

The camera imaging model is the basis for camera calibration, which establishes the geometric transformation between 3D points in the object space and 2D points in the image plane. Based on the design of spherical cameras, there are three major camera imaging models, i.e., the unified projection model, general camera model, and multi-camera model.

\begin{figure}[!t]
	\centering
	\includegraphics[width=0.48\textwidth]{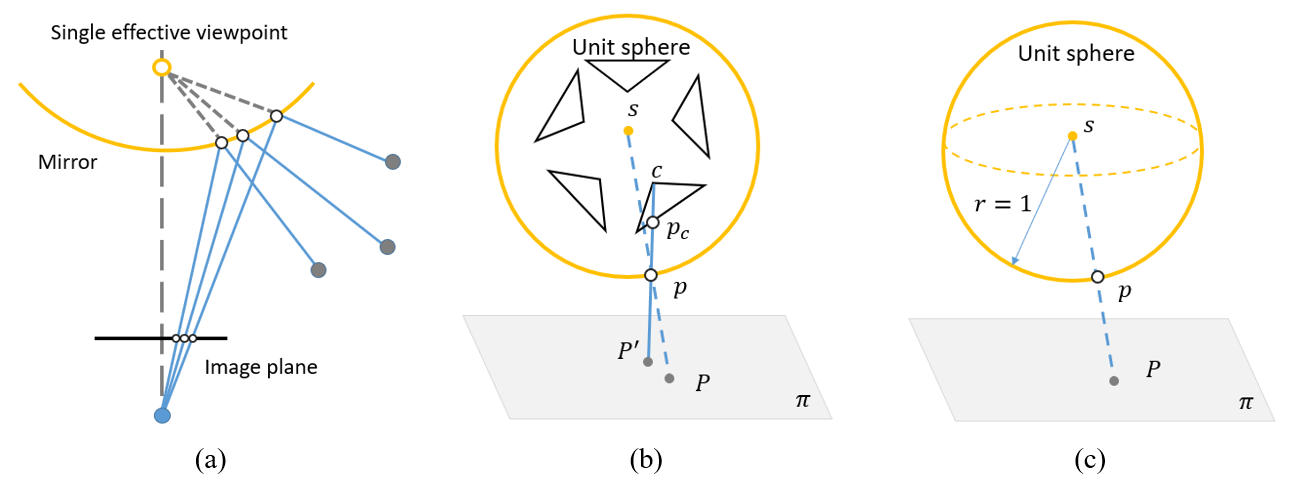}
	\caption{Camera imaging models\cite{ji2014comparison},\cite{scaramuzza2006flexible}. (a) central catadioptric cameras; (b) multi-camera model; and (c) unit sphere camera model.}
	\label{fig:figure13}
\end{figure}

\begin{itemize}
	\item \textbf{Unified projection model}\cite{mei2007single} has been mainly designed for central catadioptric cameras, in which environment light rays intersect in a single point, i.e., the projection center of the mirror, as shown in Fig. \ref{fig:figure13}(a). This camera imaging model follows a strict theoretical projection function that models real-world imaging errors. The unified projection model has recently been verified as effective for wide-angle and fisheye cameras\cite{heng2013camodocal}.
	
	\item \textbf{General camera model} (Taylor model)\cite{scaramuzza2006flexible} has been used for modeling the imaging procedure of central catadioptric and dioptric cameras. Instead of using the strict theoretical model in the unified projection model, the general camera model utilizes a Taylor polynomial function to fit the projection.
	
	\item \textbf{Multi-camera model}  has been designed to establish the projection of the widely used polydioptric cameras that record spherical images by using a camera rig, e.g., the Ladybug 5+ camera. The multi-camera model can be implemented using several individual camera models or a unit sphere camera model\cite{ji2014comparison}. The former is more rigorous in formulating the imaging system, as shown in Fig. \ref{fig:figure13}(b); on the contrary, the latter has a more straightforward formula widely used in close-range photogrammetry, as shown in Fig. \ref{fig:figure13}(c).
\end{itemize}

(2) camera calibration

The purpose of camera calibration is to calculate the intrinsic parameters of cameras, e.g., the focal length, principle point, and lens distortion coefficients. After selecting a proper camera imaging model, camera calibration can be achieved by using a combined bundle adjustment of both interior and exterior orientation parameters. In practice, there are three ways to estimate the parameters: self-calibration based on the epipolar geometry constraint, space resection using non-planar 3D control points, and laboratory calibration using checkerboards\cite{urban2015improved}. In the work of \cite{aghayari2017geometric}, a Ricoh Theta dual-camera system has been calibrated based on the expanded unit sphere camera model, in which two extra parameters that model the displacement of latitude and longitude coordinates are incorporated into the camera imaging model. For calibrating a professional Ladybug multi-camera system, \cite{lichti2020geometric} adopted the colinear equation-based rigorous model, which combines five radial distortion parameters and two decentering distortion parameters to model the imaging errors. For comparing different calibration methods for spherical cameras, readers can refer to the work of \cite{puig2012calibration}. Besides, Table \ref{tab:table4} presents a list of open-source software packages for spherical camera calibration.

\begin{table*}[t!]
	\caption{A list of open-source software packages for spherical camera calibration.}
	\label{tab:table4}
	\centering
	\makebox[\textwidth]{
		\begin{tabular}{p{0.15\textwidth} p{0.1\textwidth} p{0.1\textwidth} p{0.4\textwidth}}
			\toprule
			\textbf{Name}			& 
			\textbf{Language}       &
			\textbf{Year}           &
			\textbf{Website}		\\
			\midrule
			\multirow{2}{0.1\textwidth}{OCamCalib}
			& Matlab
			& 2006
			&\url{https://sites.google.com/site/scarabotix/ocamcalib-omnidirectional-camera-calibration-toolbox-for-matlab}\\
			& \multicolumn{3}{L{0.6\textwidth}}{A toolbox to calibrate any central omnidirectional camera, i.e., panoramic cameras having a single effective viewpoint\cite{scaramuzza2006flexible}.} \\
			
			\midrule
			\multirow{2}{0.1\textwidth}{Improved OcamCalib}
			& Matlab
			& 2015
			& \url{https://github.com/urbste/ImprovedOcamCalib}\\
			& \multicolumn{3}{L{0.6\textwidth}}{An add-on toolkit to the OCamCalib toolbox that implements calibration algorithms for wide-angle, fisheye, and omnidirectional cameras\cite{urban2015improved}.} \\
			
			\midrule
			\multirow{2}{0.1\textwidth}{LIBOMNICAL}
			& Matlab
			& 2014
			& \url{https://www.cvlibs.net/projects/omnicam}\\
			& \multicolumn{3}{L{0.6\textwidth}}{a MATLAB Toolbox to calibrate central and slightly non-central catadioptric cameras and catadioptric stereo setups\cite{schonbein2014calibrating}.} \\
			
			\midrule
			\multirow{2}{0.1\textwidth}{Omnidirectional Calibration Toolbox Mei}
			& Matlab 
			& 2007
			& \url{https://www.robots.ox.ac.uk/~cmei/Toolbox.html}\\
			& \multicolumn{3}{L{0.6\textwidth}}{A toolbox implements the unified projection model to calibrate hyperbolic, parabolic, and folding mirrors and spherical and wide-angle sensors\cite{mei2007single}.} \\
			
			\midrule
			\multirow{2}{0.1\textwidth}{camodocal}
			& C++
			& 2013
			& \url{https://github.com/hengli/camodocal}\\
			& \multicolumn{3}{L{0.6\textwidth}}{A toolbox for automatic intrinsic and extrinsic calibration of a camera rig with multiple generic cameras and odometry\cite{heng2013camodocal}.}\\
			
			\midrule
			\multirow{2}{0.1\textwidth}{kalibr}
			& C++
			& 2016
			& \url{https://github.com/ethz-asl/kalibr}\\
			& \multicolumn{3}{L{0.6\textwidth}}{A toolbox for multi-camera calibration and multi-sensor integration\cite{rehder2016extending}.} \\
			\bottomrule
		\end{tabular}
	}
\end{table*}

\subsubsection{SfM-based offline methods}
\label{sec:2.4.2}

\hfill

(1) Principle of incremental SfM

Existing SfM can be divided into three major groups, i.e., incremental SfM, global SfM, and hybrid SfM, according to the used strategy for estimating and optimizing unknown parameters\cite{cui2017hsfm}. Compared with other techniques, incremental SfM has the advantages of resisting high outlier ratios and achieving accurate orientation models. Incremental SfM has been widely used in photogrammetric 3D reconstruction\cite{jiang2017board}.

\begin{figure}[!t]
	\centering
	\includegraphics[width=0.48\textwidth]{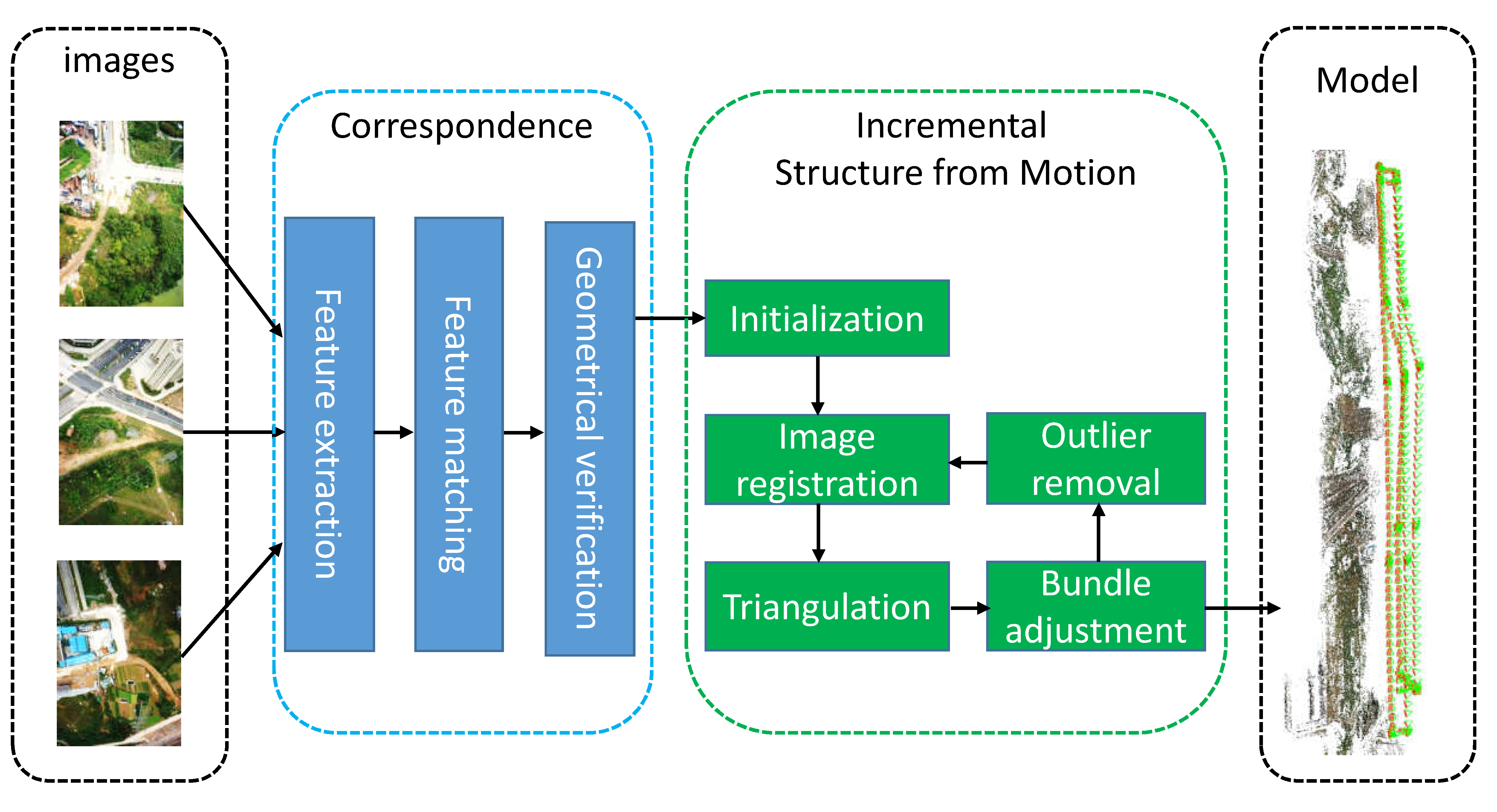}
	\caption{The workflow of the incremental SfM\cite{jiang2020effcient}.}
	\label{fig:figure14}
\end{figure}

The workflow of incremental SfM is shown in Fig. \ref{fig:figure14}, which consists of two components, i.e., feature matching and incremental image registration. Feature matching can be solved by using the methods presented in Section \ref{sec:3.1}. For image registration, consistent correspondences are first tied to create tie-points\cite{cao2019fast}. In incremental image registration, two seed images are first selected among all matched image pairs, which have a large enough intersection angle and a sufficient number of well-distributed matched features. A base model is then constructed by recovering their relative poses and triangulating 3D scene points, which would be used to register the next-best image and triangulate more 3D scene points. Meanwhile, local or global bundle adjustment (BA) optimization is executed to decrease accumulated errors and remove false matches. After iterative image registration and point triangulation, all images are registered into the same 3D model\cite{snavely2008modeling}.

In the above-mentioned iterative local and global BA, the optimization problem is usually formulated as a joint minimization of the reprojection function \cite{triggs1999bundle}, where the sum of errors between tie-point projections and their corresponding image points is minimized. The object function of BA is presented by the Equation \eqref{eq:1}.

\begin{equation}
	\min\limits_{C_j,X_i}\sum_{i=1}^n\sum_{j=1}^m\rho_{ij}\parallel P(C_j,X_i)-x_{ij} \parallel ^2
	\label{eq:1}
\end{equation}
where $X_i$ and $C_j$ indicate a 3D point and a camera, respectively; $P(C_j, X_i)$ is the projection of point $X_i$ on camera $C_j$; $x_{ij}$ is an observed image point; $\left\|\bullet\right\|$ denotes L2-norm; $\rho_{ij}$ is an indicator function with $\rho_{ij}=1$ if point $X_i$ is visible in camera $C_j$; otherwise $\rho_{ij}=0$.

(2) Perspective SfM to spherical SfM

For spherical image orientation, there have been some attempts to adapt perspective SfM to spherical SfM in the last two decades. \cite{chang2000omni} proposed a two-step SfM for omnidirectional images depending on linear initialization and nonlinear optimization, which only recovers camera relative geometry. Uncertainty analysis and result comparison were also conducted and compared with perspective SfM. In the work of \cite{torii2005two}, both two-view and three-view geometry of spherical images have been analyzed and discussed. The epipolar geometry forms the basis for 3D reconstruction from spherical images, e.g., the pose recovery in a virtual navigation system\cite{kangni2007orientation}.

\begin{figure}[!t]
	\centering
	\includegraphics[width=0.45\textwidth]{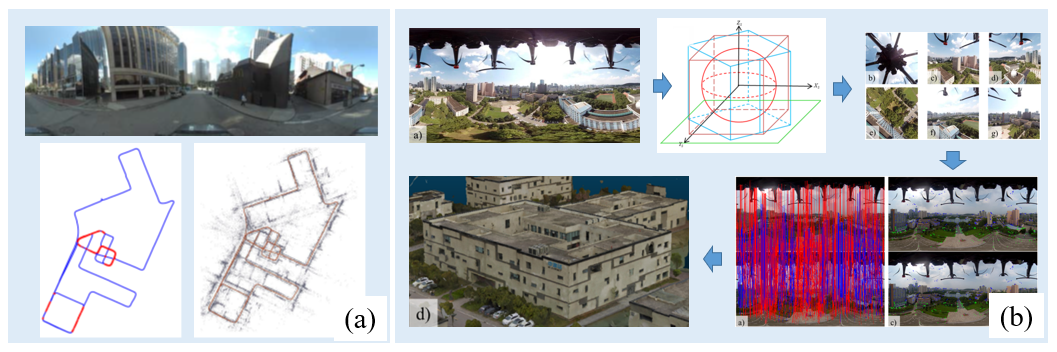}
	\caption{SfM for spherical images. (a) direct processing of spherical images based on the unit spherical camera model\cite{torii2009google}; (b) indirect processing of spherical images based on sphere-to-plane projection\cite{zhang2020uav}.}
	\label{fig:figure15}
\end{figure}

With the usage of advanced MMS systems for urban city modeling and navigation, some researchers moved their attention to 3D reconstruction of large-scale scenes instead of two or three-view geometry estimation in the earlier work\cite{micusik2006structure}. As one of the early pioneering attempts, \cite{torii2009google} proposed an incremental SfM system that combines state-of-the-art techniques, e.g., local feature-based feature detection, approximate nearest neighbor-based feature matching, and robust essential matrix estimation and optimization. The performance of the proposed SfM system has been verified by using Google Street View images that cover a large street block\cite{anguelov2010google}, as presented in Fig. \ref{fig:figure15}(a). For full spherical images, \cite{pagani2011structure} investigated different error metrics on relative and absolute pose estimation and also designed the error approximations to reduce computational costs. These error metrics consist of the basic blocks for spherical SfM.

In contrast to professional sensors, recent years also witnessed the explosive development of consumer-grade spherical cameras\cite{gao2022review} and their usage in 3D modeling. In \cite{guan2016structure}, the von Mises-Fisher distribution was utilized to model the noise distribution of feature point positions on spherical images and to reformulate the error function in the bundle adjustment optimization. Meanwhile, spherical-n-point and triangulation algorithms that are suitable for spherical images have been proposed to achieve spherical video orientation and viewing direction stabilization. \cite{zhang2020uav} embedded spherical cameras into a UAV platform and proposed a solution for panoramic photogrammetry, as presented in Fig. \ref{fig:figure15}(b). In data processing, original spherical images are first reprojected to cubic images, which are then reconstructed by using existing commercial software. The study verifies the validation of spherical cameras for oblique photogrammetric 3D modeling of urban buildings.

Although extensive research has been reported in the literature, the majority of both open-source and commercial software packages are now designed for perspective images. Table \ref{tab:table5} lists the well-known and widely used software packages for 3D reconstruction in the fields of photogrammetry and computer vision. It is shown that only a few packages provide the full 3D reconstruction module for spherical images, e.g., the open-source software OpenMVG and MicMac and the commercial software Pix4Dmapper and Metashape. With the increasing popularity of spherical cameras, there is an urgent requirement for well-designed toolkits that can support scientific research and engineering application. Thus, further research is required for photogrammetric 3D reconstruction from spherical images.

\begin{table*}[t!]
	\caption{A list of open-source and commercial SfM software packages.}
	\label{tab:table5}
	\centering
	\makebox[\textwidth]{
		\begin{tabular}{p{0.13\textwidth} p{0.1\textwidth} p{0.1\textwidth} p{0.4\textwidth} p{0.12\textwidth}}
			\toprule 
			\textbf{Name}			& 
			\textbf{Language}       &
			\textbf{Year}           &
			\textbf{Website}		&
			\textbf{Spherical camera}		\\
			\midrule
			\multirow{2}{0.1\textwidth}{AliceVision}
			& C++
			& 2018
			& \url{https://github.com/alicevision}
			& $\times$     \\
			& \multicolumn{3}{L{0.6\textwidth}}{A photogrammetric computer vision framework for 3D reconstruction. It provides three types of camera models, i.e., Radial, Brown, and Fisheye\cite{griwodz2021alicevision}.} \\
			
			\midrule
			\multirow{2}{0.1\textwidth}{COLMAP}
			& C++
			& 2016
			& \url{https://github.com/colmap/colmap}
			& $\times$     \\
			& \multicolumn{3}{L{0.6\textwidth}}{A 3D reconstruction software that includes both sparse and dense matching modules, which includes Radial, and Fisheye camera models\cite{schonberger2016structure}.} \\
			
			\midrule
			\multirow{2}{0.1\textwidth}{OpenMVG}
			& C++
			& 2015
			& \url{https://github.com/openMVG/openMVG}
			& $\checkmark$     \\
			& \multicolumn{3}{L{0.6\textwidth}}{A well-known sparse reconstruction software that provides Radial, Brown, Fisheye, and Sphere camera models. It can directly process spherical images\cite{moulon2016openmvg}.} \\
			
			\midrule
			\multirow{2}{0.1\textwidth}{MicMac}
			& C++
			& 2007
			& \url{https://github.com/micmacIGN/micmac}
			& $\checkmark$     \\
			& \multicolumn{3}{L{0.6\textwidth}}{An open-source photogrammetric toolkit that provides modules of AT, dense matching, and ortho-rectification for satellite, aerial and close-range images\cite{rupnik2017micmac}.} \\
			
			\midrule
			\multirow{2}{0.1\textwidth}{ContextCapture}
			& /
			& 2022
			& \url{https://www.bentley.com}
			& $\times$     \\
			& \multicolumn{3}{L{0.6\textwidth}}{A well-known and widely used 3D modeling software in the field of photogrammetry, which can provide the best 3D textured mesh models. It only supports perspective cameras.} \\
			
			\midrule
			\multirow{2}{0.1\textwidth}{Pix4Dmapper}
			& /
			& 2022
			& \url{https://www.pix4d.com}
			& $\checkmark$     \\
			& \multicolumn{3}{L{0.6\textwidth}}{A well-known and widely used photogrammetric software for aerial and close-range images in the field of photogrammetry. It supports perspective, fisheye, and spherical cameras.} \\
			
			\midrule
			\multirow{2}{0.1\textwidth}{Metashape}
			& /
			& 2022
			& \url{https://www.agisoft.com}
			& $\checkmark$     \\
			& \multicolumn{3}{L{0.6\textwidth}}{A well-known and widely used photogrammetric software for aerial and close-range images in the field of photogrammetry. It supports perspective, fisheye, and spherical cameras.} \\
			
			\midrule
			\multirow{2}{0.1\textwidth}{RealityCapture}
			& /
			& 2022
			& \url{https://www.capturingreality.com}
			& $\times$     \\
			& \multicolumn{3}{L{0.6\textwidth}}{A photogrammetric computer vision software for 3D reconstruction. It features a fast speed for image processing. It now supports a perspective camera.} \\
			\bottomrule
		\end{tabular}
	}
\end{table*}

\begin{table*}[t!]
	\caption{A list of open-source SLAM software packages.}
	\label{tab:table6}
	\centering
	\makebox[\textwidth]{
		\begin{tabular}{p{0.13\textwidth} p{0.1\textwidth} p{0.1\textwidth} p{0.4\textwidth} p{0.12\textwidth}}
			\toprule 
			\textbf{Name}			& 
			\textbf{Language}       &
			\textbf{Year}           &
			\textbf{Website}		&
			\textbf{Spherical camera}		\\
			\midrule
			\multirow{2}{0.1\textwidth}{ORB-SLAM3}
			& C++
			& 2021
			& \url{https://github.com/UZ-SLAMLab/ORB_SLAM3}
			& $\times$     \\
			& \multicolumn{3}{L{0.6\textwidth}}{A SLAM system that supports perspective and fisheye cameras and can be embedded with monocular, stereo, and RGBD cameras\cite{campos2021orb}.} \\
			
			\midrule
			\multirow{2}{0.1\textwidth}{OpenVSLAM}
			& C++
			& 2019
			& \url{https://github.com/xdspacelab/openvslam}
			& $\checkmark$     \\
			& \multicolumn{3}{L{0.6\textwidth}}{A visual SLAM system that supports various camera models, including perspective, fisheye, and spherical cameras\cite{sumikura2019openvslam}.} \\
			
			\midrule
			\multirow{2}{0.1\textwidth}{Cubemap-SLAM}
			& C++
			& 2018
			& \url{https://github.com/nkwangyh/CubemapSLAM}
			& $\times$     \\
			& \multicolumn{3}{L{0.6\textwidth}}{A visual SLAM system that converts fisheye images into perspective images and achieves image orientation based on the ORB-SLAM\cite{wang2018cubemapslam}.} \\
			
			\midrule
			\multirow{2}{0.1\textwidth}{DSO}
			& C++
			& 2017
			& \url{https://github.com/JakobEngel/dso}
			& $\times$     \\
			& \multicolumn{3}{L{0.6\textwidth}}{A direct sparse visual odometry that supports only perspective cameras\cite{engel2017direct}.} \\
			
			\midrule
			\multirow{2}{0.1\textwidth}{MultiCol-SLAM}
			& C++
			& 2016
			& \url{https://github.com/urbste/MultiCol-SLAM}
			& $\times$     \\
			& \multicolumn{3}{L{0.6\textwidth}}{A multi-fisheye SLAM that supports rigidly coupled multi-camera systems. It extends the ORB-SLAM and ORB-SLAM2 systems\cite{urban2016multicol}.} \\
			
			\midrule
			\multirow{2}{0.1\textwidth}{LSD-SLAM}
			& C++
			& 2014
			& \url{https://github.com/tum-vision/lsd_slam}
			& $\times$     \\
			& \multicolumn{3}{L{0.6\textwidth}}{A direct monocular SLAM system that enables real-time image orientation and semi-dense depth map generation\cite{engel2014lsd}.} \\
			\bottomrule
		\end{tabular}
	}
\end{table*}

\subsubsection{SLAM-based online methods}
\label{sec:2.4.3}

\hfill

(1) Principle of visual SLAM

In contrast to the offline SfM technique, SLAM can implement simultaneous and real-time image orientation and scene reconstruction, which origins from the robotic field for localization and navigation without GNSS signals, as well as environment mapping\cite{huang2019survey}. In the field of photogrammetry and computer vision, SLAM has also been used for online image orientation, e.g., UAV and MMS images\cite{huang2020fast},\cite{lu2018survey}. Various sensors can be integrated into a SLAM system, such as RGB and depth cameras, laser scanners, GNSS and IMU (Inertial Measurement Unit) instruments, and wheel odometry\cite{huang2019survey}. For spherical image orientation, the used SLAM system is termed visual SLAM.

\begin{figure}[!t]
	\centering
	\includegraphics[width=0.48\textwidth]{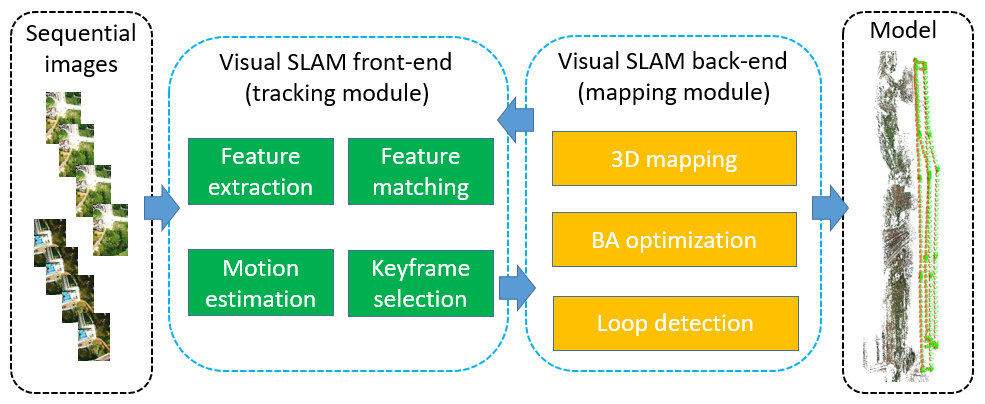}
	\caption{The workflow of the visual SLAM system.}
	\label{fig:figure16}
\end{figure}

Fig. \ref{fig:figure16} shows the workflow of the visual SLAM, which consists of two major components. The front-end is utilized to process sequentially observed images through feature extraction, feature matching, motion estimation, and keyframe selection; the back-end is responsible for loop detection, BA optimization, and 3D mapping. The processing pipeline of the visual SLAM system includes: 1) sequentially estimating the motion of newly added images with local BA optimization; 2) checking whether or not the newly added images are keyframes and creating more map points from keyframes; 3) detecting loops between newly added images and existing 3D models and conducting global BA optimization to reduce error accumulation. Through the iterative execution of these three steps, images are sequentially oriented.

(2) SLAM with large FOV cameras

Cameras with large field-of-view angles can enhance the robustness of the SLAM systems\cite{zhang2016benefit} since they enable long-period feature tracking, especially in urban streets and indoor environments. Large FOV cameras have been increasingly integrated into visual SLAM systems. Recently adopted large FOV cameras can be grouped into two categories, including fisheye cameras with wide FOV and spherical cameras with full FOV.

Fisheye cameras are the most reported large FOV sensors coupled with visual SLAM systems\cite{caruso2015large,matsuki2018omnidirectional,won2020omnislam,zhang2016benefit}. \cite{caruso2015large} proposed a direct monocular SLAM based on a unified omnidirectional camera model. It was an extension of their previous LSD-SLAM\cite{engel2014lsd}, and it has superior performance on the accuracy of localization and robustness to strong rotational movement. In the work of \cite{matsuki2018omnidirectional}, the authors also designed a direct monocular SLAM system, which is an extension of the DSO system\cite{engel2017direct} and integrates the unified camera model proposed in \cite{caruso2015large}. \cite{wang2018cubemapslam} proposed a feature-based SLAM system, termed Cubemap-SLAM, based on the cubic projection of fisheye images. In this work, the cubemap model has been embedded into the ORB-SLAM. To evaluate the impact of large FOV cameras, \cite{zhang2016benefit} conducted a series of tests by using both indoor and outdoor datasets, and their evaluation reveals that the reconstruction accuracy also depends on the environment, and large FOV cameras tend to improve performance in indoor scenes.

\begin{figure}[!t]
	\centering
	\includegraphics[width=0.48\textwidth]{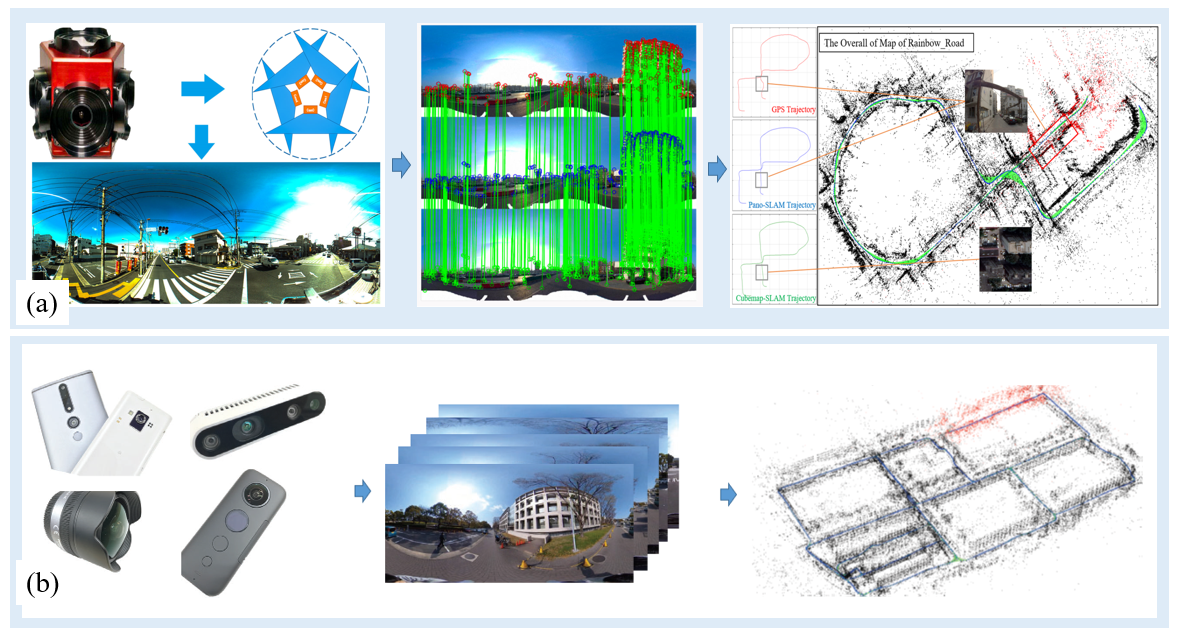}
	\caption{The visual SLAM systems based on (a) the multi-fisheye camera rig\cite{ji2020panoramic} and (b) the full spherical camera\cite{sumikura2019openvslam}.}
	\label{fig:figure17}
\end{figure}

Spherical cameras can further extend the FOV of fisheye cameras, and they have been widely used for data acquisition in MMS systems\cite{anguelov2010google,torii2009google}. \cite{tardif2008monocular} proposed a feature-based SLAM workflow that only estimates the pose of the latest image without the execution of BA optimization. The core idea is to decouple the problem of pose recovery into rotation estimation from epipolar geometry and translation estimation using 3D-2D correspondences. \cite{urban2016multicol} extended the ORB-SLAM\cite{mur2015orb,mur2017orb} system into a multi-fisheye omnidirectional SLAM system, termed MultiCol-SLAM, which supports the arbitrary rigidly coupled multi-camera systems. In \cite{ji2020panoramic}, a fisheye calibration model has been designed and used for multiple fisheye camera rigs, as shown in Fig. \ref{fig:figure17}(a), and a feature-based SLAM, termed PAN-SLAM, has been proposed based on well-designed strategies for initialization, feature matching and tracking, and loop detection. The tests demonstrate the validation of the proposed SLAM system.

In recent years, spherical images captured by dual-fisheye cameras are becoming more and more popular, which promotes the development of panoramic SLAM systems\cite{huang2022360vo,im2016all,sumikura2019openvslam,zhang2021panoramic}. \cite{sumikura2019openvslam} designed a versatile visual SLAM, termed OpenVSLAM, which implements perspective, fisheye, and spherical camera models and supports monocular, stereo, and RGB-D cameras, as shown in Fig. \ref{fig:figure17}(b). In addition, the authors provided an indoor benchmark for performance evaluation\cite{chappellet2021benchmarking}. \cite{zhang2021panoramic} investigated varying feature detection and description techniques, including SPHORB and SSIFT, and compared their performance in spherical image orientation based on SLAM. Recently, \cite{huang2022360vo} implemented a direct SLAM system that extends DSO (Direct Sparse Odometry) with the spherical camera model to process equirectangular images. Table \ref{tab:table6} presents a list of open-source SLAM software packages. The same finding can be made as for SfM-based methods that few SLAM systems support full spherical cameras.

\subsection{Dense matching}
\label{sec:2.5}

Dense matching aims to establish pixel-wise correspondences and generate point clouds from SfM or SLAM-based oriented images. According to the 3D reconstruction pipeline shown in Fig. \ref{fig:figure2}, generated dense point clouds would be used to construct detailed 3D models after point meshing and texture mapping. Thus, the performance of dense matching methods would determine the precision and completeness of the final 3D models. In the literature, dense matching has been an extensively studied topic with the arising of 3D reconstruction in photogrammetry and computer vision. In addition to traditional methods for perspective images, recent years have witnessed increasingly reported dense matching methods for spherical images. Generally, this work can be grouped into single-view depth prediction and multi-view stereo matching.

%Table \ref{tab:table7} lists the methods for dense matching.

%\begin{table*}[t!]
%	\centering
%	% \footnotesize
%	\caption{Method comparison for dense matching.}
%	\label{tab:table7}
%	\makebox[\linewidth]{
%		\begin{tabular}{p{0.14\textwidth} p{0.15\textwidth} p{0.45\textwidth}}
%			\toprule
%			\textbf{Category} &	\textbf{Methods} & \textbf{Advantages and disadvantages} \\
%			\midrule
%			Single-view prediction & Single view	\cite{feng2022360,jang2022egocentric,jin2020geometric,pintore2021slicenet,zioulis2018omnidepth}
%			& \textbf{\emph{Advantages}}: (1) avoid SfM and SLAM-based image orientation.
%			\textbf{\emph{Disadvantages}}: (1) do not use multi-view constraints. \\
%			\midrule
%			Existing methods & Hand-crafted methods\cite{furukawa2010accurate,hirschmuller2007stereo,rothermel2012sure} and learning-based methods\cite{chang2018pyramid,seki2017sgm,yao2020blendedmvs}
%			& \textbf{\emph{Advantages}}: (1) use existing algorithms and software packages; (2) be suitable for engineer usage.
%			\textbf{\emph{Disadvantages}}: (1) cannot avoid the distortions or require extra time costs for image projection. \\
%			\midrule
%			Redesigned methods & Stereo-view methods\cite{meuleman2021real,wang2020360sd} and multi-view methods\cite{jang2022egocentric}
%			& \textbf{\emph{Advantages}}: (1) adapt to spherical images; (2) make use of the learning-based CNN network.
%			\textbf{\emph{Disadvantages}}: (1) design special acquisition sensors and (2) label data for network training. \\
%			\bottomrule
%		\end{tabular}
%	}
%\end{table*}

\subsubsection{Single-view depth prediction}
\label{sec:2.5.1}

\begin{figure}[!t]
	\centering
	\includegraphics[width=0.45\textwidth]{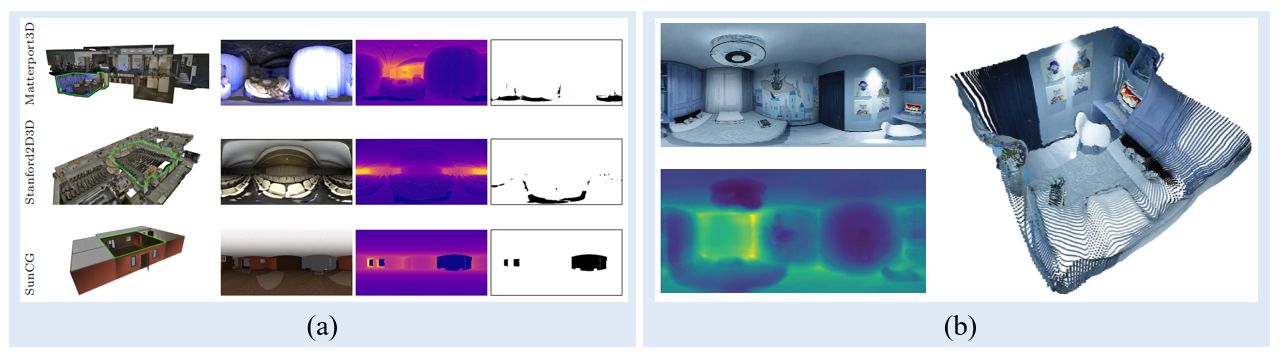}
	\caption{Single view depth prediction for spherical images. (a) a prepared 360 dataset\cite{zioulis2018omnidepth}; (b) indoor depth prediction from spherical images\cite{jin2020geometric}.}
	\label{fig:figure18}
\end{figure}

Single-view depth prediction methods can process an individual spherical image instead of pixel-wise correspondence searching between images. In the literature, single-view depth prediction methods are usually implemented through learning-based CNN networks\cite{feng2022360,jang2022egocentric,jin2020geometric,pintore2021slicenet,zioulis2018omnidepth}, which belong to a new research topic in the last five years. As one of the earliest works, \cite{zioulis2018omnidepth} prepared and released a spherical training dataset with ground-truth depth maps, as presented in Fig. \ref{fig:figure18}(a). It can be used as the training dataset for depth estimation networks for spherical images instead of training on perspective datasets with sub-optimal performance. Considering the correlation between depth and geometric structure in indoor environments, \cite{jin2020geometric} attempted to leverage existing geometric structures, e.g., corners, boundaries, and planes, as priors for network training or inferring these structures from generated depth estimation. The reconstruction models are presented in Fig. \ref{fig:figure18}(b). For indoor scenes, \cite{pintore2021slicenet} suggested representing equirectangular images by using vertical slices of the sphere, which partitions input images into vertical slices. Based on the LSTM (long short-term memory), the authors designed a network called SliceNet for depth estimation.

In contrast to indoor scenes, \cite{feng2022360} prepared the Depth360 dataset and designed an end-to-end two-branch network, termed SegFuse, for depth prediction of spherical images. The core idea is to integrate the segmentation of cubic-map images from one branch into the depth estimation of equirectangular images from the other branch.

\begin{figure}[!t]
	\centering
	\includegraphics[width=0.45\textwidth]{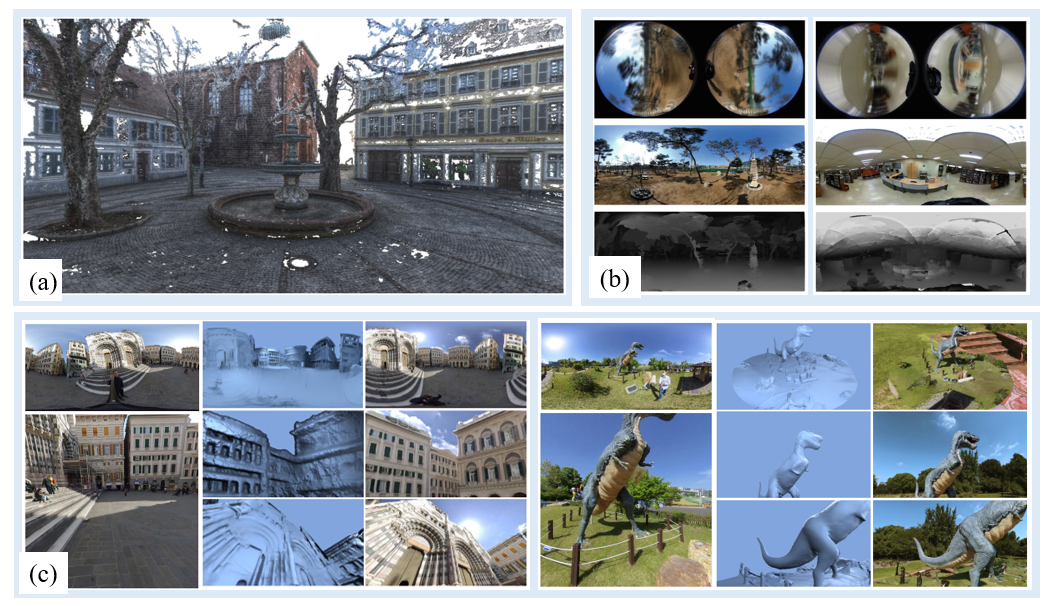}
	\caption{multi-view dense matching for spherical images. (a) revised PMVS algorithm\cite{pagani2011dense}; (b) sphere sweeping algorithm\cite{im2016all}; and (c) the finetuned CNN network\cite{jang2022egocentric}.}
	\label{fig:figure19}
\end{figure}

\subsubsection{Multi-view stereo matching}
\label{sec:2.5.2}

\hfill

(1) existing method

Multi-view stereo matching methods attempt to recover dense point clouds from two or multiple oriented images based on SfM and SLAM image orientation techniques\cite{furukawa2015multi}. Existing methods can be divided into two groups based on the strategies used. For the first one, traditional hand-crafted methods are on the one hand revised to adapt to spherical images. In the work of\cite{pagani2011structure,pagani2011dense}, the PMVS (Patch-based Multi-view Stereo) library was revised to directly process spherical images for dense matching, and the results are shown in Fig. \ref{fig:figure19}(a). For the second one, spherical images are projected into cubic-map images, and traditional hand-crafted algorithms and deep learning-based networks are adopted straightforward since they are designed or trained for perspective images, e.g., hand-crafted methods\cite{bleyer2011patchmatch,furukawa2010accurate,hirschmuller2007stereo,rothermel2012sure} and learning-based methods\cite{chang2018pyramid,khamis2018stereonet,seki2017sgm,yao2018mvsnet,yao2020blendedmvs}. These two strategies can make use of existing well-designed methods.

(2) redesigned method

For this category, algorithms are designed and implemented from scratch considering the characteristic of spherical images\cite{da2019dense,im2016all,kim20133d,meuleman2021real,wang2020360sd}. In \cite{kim20133d}, a full workflow was designed for 3D reconstruction of spherical image pairs, in which a stereo matching algorithm was proposed based on a partial differential equation (PDE), and the complete 3D scene was obtained by registering partial models. Considering the power of the plane sweeping algorithm, \cite{im2016all} proposed a sphere sweeping algorithm based on the unified omnidirectional camera model, in which virtual spheres were used instead of virtual planes. Matching results are shown in Fig. \ref{fig:figure19}(b). The algorithm was tested by using both synthetic and real datasets. These methods can be seen as mimics of the traditional dense matching algorithm for perspective images.

Recently, CNN networks have also been adopted to design multi-view stereo solutions. In the work of \cite{wang2020360sd}, a dual-camera imaging system that consists of top and bottom cameras was designed, which ensures that the epipolar lines of captured images are vertically aligned. A polar angle layer was added to a two-branch network for depth estimation, which plays as additional input geometric information to supervise model training. \cite{jang2022egocentric} proposed a complete solution for 3D reconstruction from spherical images, including image orientation, epipolar rectification, dense matching, and texture mapping. For dense matching, the authors adopted an existing network that was retrained using a synthetic dataset to learn spherical disparity and designed a spherical binoctree structure for depth map fusion. The proposed solution achieves superior performance when compared with traditional methods, as presented in Fig. \ref{fig:figure19}(c). In conclusion, dense matching methods based on CNN networks attract attention in recent years. This review would not cover all aspects in this field.

\section{Applications}
\label{sec:3}

This section presents promising applications related to spherical images. Due to the characteristics of full FOV and low cost, spherical images have been used widely in a variety of applications. According to the purpose of this study, the reviewed applications would be restricted to 3D reconstruction, including cultural heritage documentation, urban modeling and navigation, tunnel mapping and inspection, and other applications, e.g., urban tree localization, emergency response and rescue, and underwater occlusion avoidance.

\subsection{Cultural heritage documentation}
\label{sec:3.1}

Cultural heritage documentation (CHD) is the earliest usage of spherical images, which can be dated back to two centuries ago\cite{luhmann2004historical}. In contrast to the large-scale acquisition required in aerial photogrammetry, the scale of cultural heritage documentation is much smaller but with serious occlusions, e.g., building pillars and inner structures. It requires close-range data acquisition with a large FOV to decrease human labor and ensure completeness. When compared to conventional aerial plane-based and recent UAV-based photogrammetry, spherical images can be recorded in a more flexible way, e.g., ground-fixed tripods and hand-held poles, and have been widely used for cultural heritage documentation.

In the field of photogrammetry and remote sensing,\cite{fangi2007multi,fangi2010multiscale,fangi2013photogrammetric} have made an earlier contribution to CHD by using spherical images. This work presents the basic principle of the proposed spherical photogrammetry (SP), i.e., the collinear equation of spherical imaging and the coplanarity condition for image orientation. By using two church architectures, this study verifies the advantages of spherical images with the remarkable efficiency of data acquisition, the completeness of documentation, and the low economic cost compared with other photogrammetric instruments.

With the explosive development of consumer-grade spherical cameras, recent researchers have also turned to adopting low-cost cameras for data acquisition and 3D reconstruction of CHD. For example, \cite{fangi2018improving} used a Panono 360 camera for the image collection of two churches, which is a ball-shaped camera that consists of 36 camera submodules to record surrounding environments. This study verified the centimeter-level precision of dense point clouds from SfM and MVS-based photogrammetric processing. \cite{barazzetti2018can} exploited a Xiaomi Mi Sphere 360 camera for the precision assessment of 3D modeling. In this study, a Leica TS30 total station has been used for an in-site survey of ground control points (GCPs), and two software packages, including Agisoft Metashape and Pix4dMapper, have been evaluated. Experiments demonstrate that millimeter-level accuracy has been obtained in both image orientation and 3D reconstruction by using GCPs and laser scanning point clouds.

\begin{figure}[!t]
	\centering
	\includegraphics[width=0.48\textwidth]{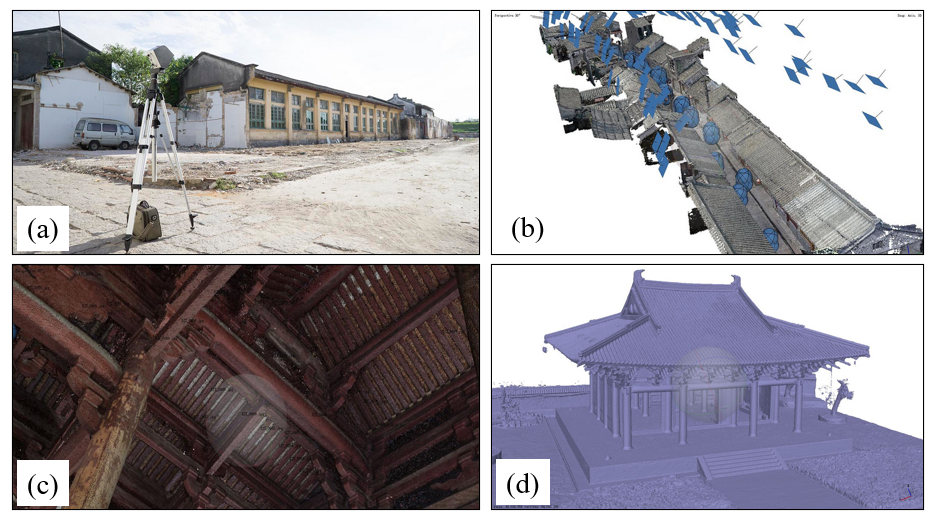}
	\caption{Cultural heritage documentation by using aerial UAV images and terrestrial spherical images\cite{2022Agisoft}. (a) Data acquisition based on the Weiss AG Civetta camera; (b) image orientation of aerial-ground images; (c) dense matching of building inner structures; (d) 3D reconstruction of buildings.}
	\label{fig:figure20}
\end{figure}

Because of the complex structure of cultural heritages, aerial and ground images have also been combined to reconstruct complex cultural heritages. \cite{herban2022use} combined low-altitude UAV images and hand-held spherical images for 3D documentation of a bell tower. In this work, a DJI Phantom 4 Pro UAV has been adopted for scanning external structures under a properly designed flight trajectory, and two spherical cameras, i.e., the GoPro Fusion and the Kandao Qoocam 8K, have been used for scanning internal structures. Based on an SfM and MVS-based processing solution, the reconstructed 3D model has high-quality details in the external and internal of the complex bell tower. In contrast to using consumer-grade spherical cameras, \cite{2022Agisoft} conducted high-quality digital twin documentation of a well-known town, Long Hu Gu Zhai at Guangdong Province, China. This study uses professional Weiss AG Civetta spherical cameras for terrestrial images with 230 megapixels resolution and a UAV for along-street aerial images. By using the Agisoft Metashape, 3D models were reconstructed from aerial and terrestrial images, as presented in Fig. \ref{fig:figure20}. The results demonstrate the potential of aerial and ground images for the precision documentation of complex buildings.

\subsection{Urban modeling and navigation}
\label{sec:3.2}

\begin{figure}[!t]
	\centering
	\includegraphics[width=0.48\textwidth]{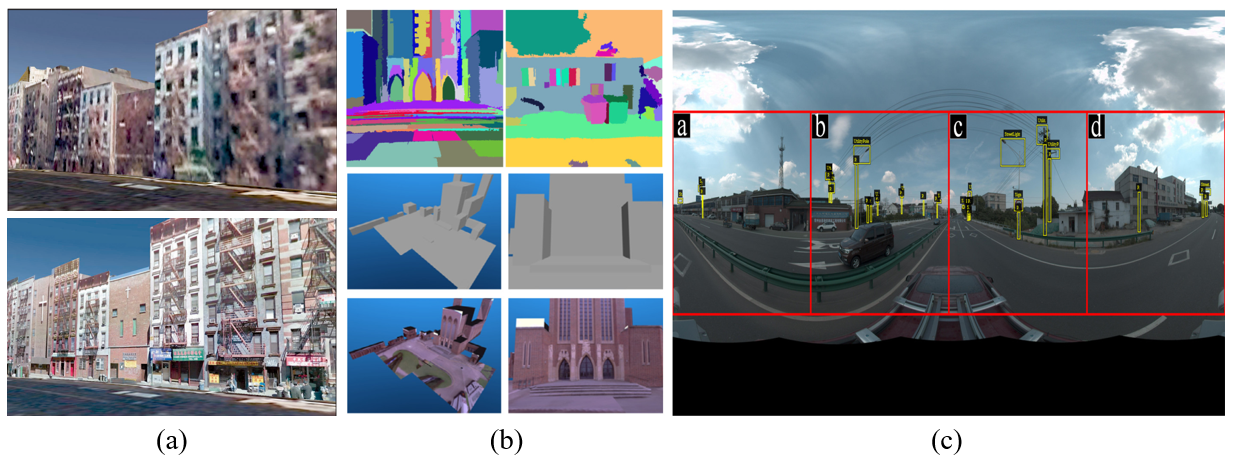}
	\caption{Urban modeling and navigation. (a) mesh models from Google Street View\cite{anguelov2010google}; (b) 3D models from spherical images\cite{kim20133d}; (c) street object localization\cite{li2022optimized}.}
	\label{fig:figure21}
\end{figure}

Spherical cameras have been integrated with mobile mapping systems (MMS) for urban street modeling and navigation. The most famous application of spherical images comes from Google Street View, which provides immersive navigation and observation along urban streets. At the beginning of Google Street View, collected street images have also been used for 3D modeling of building facades in addition to image navigation\cite{anguelov2010google},\cite{bruno2019accuracy}, as shown in Fig. \ref{fig:figure21}(a). \cite{micusik2009piecewise} designed a unified framework to reconstruct 3D models from spherical images captured from a Ladybug camera, in which bundle adjustment free image orientation, piecewise planarity constrained dense matching, and novel depth map fusion were reported considering the characteristics of low-texture, repetitive pattern, and large light changes in urban streets.

In contrast to mesh-based 3D models, recent studies show well-formed wire-frame models reconstructed from spherical images. Under the Manhattan world assumption, \cite{kim20133d} converted spherical images into cubic-map representation and segmented urban scenes into planar structures under the constraints of image color, and edge and normal information from MVS-based depth maps, as presented in Fig. \ref{fig:figure21}(b). By using street images and deep learning techniques, \cite{xu2022building} designed a workflow for building detection and height calculation and created 3D models for large-scale urban scenes.

Due to the high image overlap of large FOV cameras, spherical images have also been widely used to achieve localization and navigation in complex urban environments. \cite{cheng2018crowd} proposed a solution for the large-scale localization of photos without geotagged labels. In this study, 3D sparse models reconstructed from street view images were used as the reference data source, and a three-step algorithm was designed for geo-localization, which includes image retrieval-based coarse geo-localization, reliable feature matching between the query image and retrieved candidate images, and the PnP (Perspective n Points) based precise geo-localization. Instead of using low-level features, \cite{jayasuriya2020active} turned to detect high-level semantic information from spherical images, such as lamp posts and street signs, which are detected based on a retrained YOLO CNN network. To geo-localizing interesting targets within urban scenes to assist vehicle navigation, \cite{li2022optimized} proposed a line of bearing (LOB) based positioning method for urban street objects, e.g., road lamps, shown in Fig. \ref{fig:figure21}(c).

\subsection{Tunnel mapping and inspection}
\label{sec:3.3}

Artificial tunnels are extensively used in daily life, e.g., traffic tunnels and drainage pipes. Regular inspection is essential to ensure normal functions and extend the lifespans. In practice, the inspection work has been widely achieved through labor-sensitive operators. In recent years, with the advent of small-size spherical cameras and the development of image processing techniques, image-based inspection solutions have also been presented and verified to achieve automatic tunnel mapping and inspection\cite{leingartner2016evaluation}.

\begin{figure}[!t]
	\centering
	\includegraphics[width=0.48\textwidth]{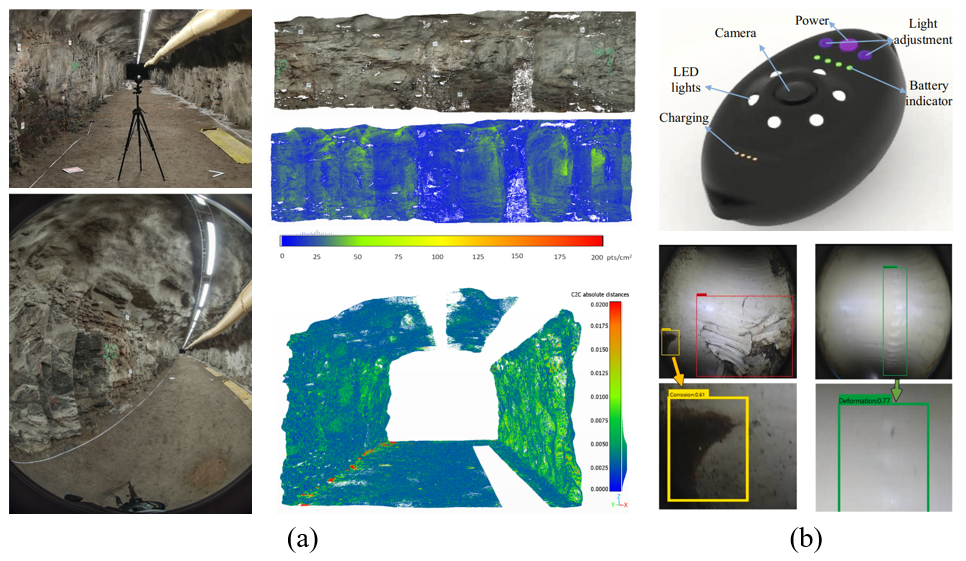}
	\caption{Tunnel mapping and inspection. (a) mine tunnel mapping\cite{janiszewski2022rapid}; (b) drainage pipe inspection\cite{guo2022detection}.}
	\label{fig:figure22}
\end{figure}

In the work of \cite{zhu2016panoramic}, a low-cost and flexible system has been designed to create a panorama image of the traffic tunnels, in which SfM-based image orientation and stitching-based image registration techniques have been used under the constraint of tunnel design data. This work verified the feasibility of image-based mapping for tunnel in-field inspection. Based on the rapid data acquisition ability of spherical cameras, \cite{janiszewski2022rapid} designed an overall framework for photogrammetric tunnel mapping. In this work, an Insta360 Pro sphere camera integrating six fisheye camera modules has been used to record images with 4000 by 3000 pixels. During data acquisition, the spherical camera was installed on a tripod to image recording with long exposure times. For quality verification, point clouds were also collected by using a Riegl VZ-400i TLS scanner. Experimental results demonstrate that the reconstructed tunnel models can achieve millimeter level precision when compared with TLS point clouds, as presented in Fig. \ref{fig:figure22}(a).

In contrast to large-size tunnel mapping, small-size gas and drainage pipes are the other typical inspection scenes that require the full FOV acquisition ability of a spherical camera due to the limited space in these pipes. In the work of\cite{zhang20193d}, drainage pipe inspection has been implemented by using low-cost GoPro Fusion video cameras. Key frames were extracted from recorded videos and converted to cubic-map images, then used to reconstruct 3D models based on typical SfM and MVS-based solutions. For the inspection of drainage pipes with minimal space, \cite{fang2022sewer} designed a pipeline capsule machine (PCM), as presented in Fig. \ref{fig:figure22}(b), which mainly consists of a large FOV camera, a LED (light-emitting diode) lighting instrument, and a power supply module. The PCM sensor can record images with the flow of water, which are further used for crack and erosion detection based on deep learning-based methods\cite{guo2022detection}.

Similarly, for the inspection of gas pipes,\cite{karkoub2020gas} designed an inspection robot that is equipped with a catadioptric camera for spherical image collection. Unlike the PCM, depending on flowing water, the inspection robot can operate on other pipes since it uses wheels for forward moving. In a word, spherical cameras have good advantages for tunnel mapping and inspection due to the low cost of sensors and the limited space of tunnels.

\subsection{Other applications}
\label{sec:3.4}

In recent years, spherical images have also been used in other applications, e.g., emergency response and rescue, urban tree detection and mapping, collision avoidance of underwater vehicles, etc. \cite{jhan2021integrating} designed an integrated photogrammetric system for rapid data acquisition and damaged building inspection. This work took advantage of low-altitude UAVs and ground backpack MMSs to collect aerial-ground images and validate the integration of aerial-ground images for the full 3D reconstruction of urban buildings to boost the observation completeness.

\begin{figure}[!t]
	\centering
	\includegraphics[width=0.48\textwidth]{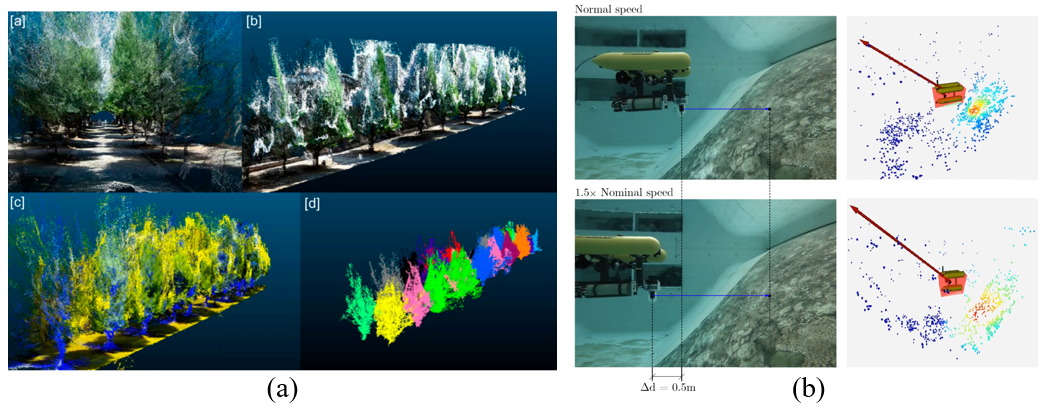}
	\caption{Spherical cameras for (a) urban tree detection\cite{itakura2020automatic} and (b) obstacle avoidance of underwater vehicles\cite{ochoa2022collision}.}
	\label{fig:figure23}
\end{figure}

Tree number, location, and structure are critical for the precision management of vegetation resources. Although conventional aerial photogrammetry and LiDAR (Light Detection and Ranging) can provide large-scale RS data, they suffer serious occlusions of urban scenes and cannot achieve accurate measurements of street trees. Considering these issues, \cite{itakura2020automatic} proposed using spherical cameras for the 3D modeling and parameter calculation of urban trees. In this study, 3D models were reconstructed based on SfM, which were further processed to extract individual trees, as shown in Fig. \ref{fig:figure23}(a). Based on the 3D models, some useful parameters, such as trunk diameter and tree height, can be then calculated. Similarly, \cite{lumnitz2021mapping} adopted a deep learning network for individual tree detection from urban street images and depended on the single-view depth estimation and triangulation techniques to calculate the precise tree geo-localization.

The underwater image collection is a frequent task for marine exploration, which has been completed by using autonomous underwater vehicles. Acoustic sensors are widely utilized instruments for obstacle avoidance. However, they cannot work well at a close range. In the work of \cite{ochoa2022collision}, an obstacle detection instrument has been designed by using a spherical camera, which can generate real-time point clouds from a visual SLAM system and assist the autonomous obstacle detection, as presented in Fig. \ref{fig:figure23}(b).

\section{Prospects}
\label{sec:4}

3D reconstruction from spherical images has been greatly promoted by the development of professional and consumer-grade cameras and automatic image processing techniques, as well as their increasing usage in varying fields. Compared with classical perspective images, there still exist some challenges for 3D reconstruction based on spherical images, which can be categorized as exploitation of crowdsource image, spatial resolution of spherical image, camera distortion and calibration, and integration of aerial-ground image. The details are presented in the following subsections.

\subsection{Exploitation of crowdsource image}
\label{sec:4.1}

\begin{figure}[!t]
	\centering
	\includegraphics[width=0.45\textwidth]{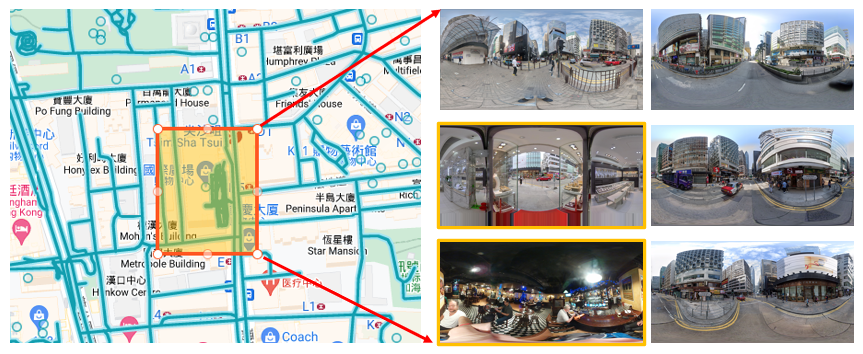}
	\caption{Crowdsource images from Google Street Views.}
	\label{fig:figure24}
\end{figure}

In recent years, spherical images can be easily recorded based on low-cost and easy-to-use spherical cameras, as presented in Fig. \ref{fig:figure4}. Besides, spherical images can also be downloaded from well-known map providers, e.g., Google Maps and Tencent Maps. These crowdsource images are usually freely provided for non-commercial usage, which plays a critical role in spherical image-based applications\cite{biljecki2021street}. However, 3D reconstruction based on these crowdsource images is non-trivial for two main reasons. On the one hand, the diversity of crowdsource images is very large, which can be recorded by using professional or consumer-grade cameras, and obtained from indoor or outdoor scenes, as shown in Fig. \ref{fig:figure24}; on the other hand, the overlap degree of crowdsource images can not be ensured since they are captured by non-professional users or just for the visual navigation purpose. These factors can frequently cause the failure of 3D reconstruction.

Crowdsource images still have great potential in 3D reconstruction, especially for urban buildings, although they have the above-mentioned drawbacks. There are some possible solutions to exploit the potential of freely available crowdsource images. First, image classification methods can be used to separate indoor and outdoor images, e.g., recent deep learning-based networks. The classified images can be used for subsequent indoor or outdoor 3D modeling. Second, existing crowdsource images can be useful compensation from the street viewpoints, such as for UAV images from aerial viewpoints. In this situation, crowdsource images could be registered to the reconstructed 3D models from UAV images, avoiding the requirement of high overlap degrees in image-based 3D reconstruction.

\subsection{Spatial resolution of spherical image}
\label{sec:4.2}

Spherical cameras record the 360 by 180 degrees of surrounding environments using only one shot, which dramatically accelerates data acquisition. However, because of the limited CCD (Charge Coupled Device) dimensions of cameras, the spatial resolution of collected images is also obviously decreased when compared with classical perspective cameras with images of the same CCD dimensions. The spatial resolution causes a contradiction between the efficiency of data acquisition and the quality of the reconstructed model.

\begin{itemize}
	\item \textbf{The efficiency of data acquisitions.} Much more time would be consumed to record images with higher resolutions to ensure the enough camera exposure time. For example, a Weiss AG Civetta spherical camera can record images with a dimension of 230 M pixels as it uses a rotation lens to scan surrounding scenes. This camera, however, can only be installed on a ground-fixed tripod since it consumes 40 seconds for each camera exposure.
	
	\item \textbf{The quality of reconstructed models.} The consumer-grade cameras usually consist of two or more dioptric camera modules, such as the Theta X camera with two lenses. This design model decreases the time costs consumed in outdoor data acquisition. However, they sacrifice the spatial resolution of recorded images, which can further reduce the quality of reconstructed models.

\end{itemize}

In order to increase the spatial resolution of spherical images, two possible solutions can be implemented without sacrificing the efficiency of data acquisition. On the one hand, more cameras can be integrated into one spherical camera as the spherical image is the fusion of images from these camera modules. This strategy has been widely used in recent spherical cameras, such as the Ladybug 5+ with six cameras and the Panono with 36 cameras; on the other hand, high-speed imaging could be a promising technology. It does not increase the size and weight of spherical cameras, and it decreases the time costs for recording high-resolution images\cite{feng2016measurement}.

\subsection{Camera distortion and calibration}
\label{sec:4.3}

\begin{figure}[!t]
	\centering
	\includegraphics[width=0.48\textwidth]{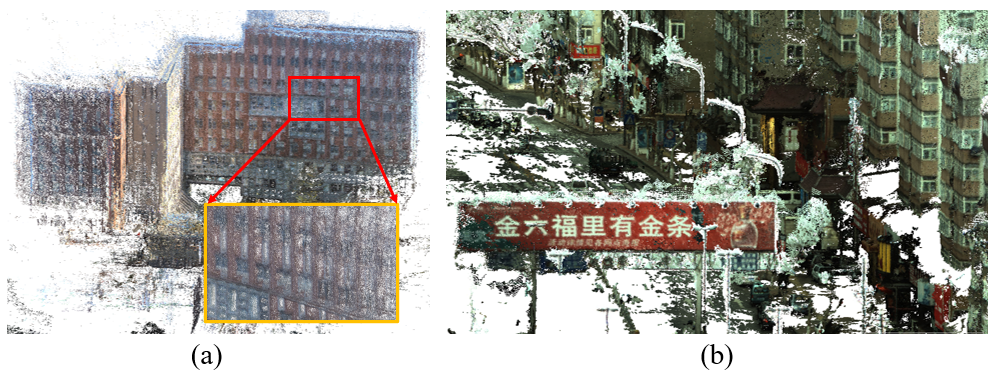}
	\caption{The illustration of dense matching point clouds from (a) a consumer-grade camera; and (b) a professional camera.}
	\label{fig:figure25}
\end{figure}

Serious geometric distortions exist in spherical images. On the one hand, the individual camera module has imaging geometric distortions; on the other hand, generating the spherical image would also introduce distortions. Especially for consumer-grade cameras, such as Ricoh Theta X and Insta360 Sphere, these cameras project multiple images from each camera onto a sphere and generate the ERP format spherical images.
Generally, an ideal unit sphere camera model is used in the generation of the ERP format, as illustrated in Fig. \ref{fig:figure13}(c), including the well-known commercial and open-source software packages Agisoft Metashape, Pix4Dmapper, and OpenMVG. This projection would introduce distortions and cause errors in 3D reconstruction. Fig. \ref{fig:figure25} presents a comparison of 3D models reconstructed from consumer-grade and professional spherical images. It is clearly shown that the point clouds in Fig. \ref{fig:figure25}(a) are very coarse, even on the plain wall. On the contrary, the point clouds generated using the professional camera are more accurate, from which the text is very clear. Therefore, camera calibration should be seriously considered and executed for spherical images to improve the quality of reconstructed models.

\subsection{Integration of aerial-ground image}
\label{sec:4.4}

Although spherical cameras can record 360 by 180 degrees of surrounding environments, their effective observation range is still limited to the near-ground region. In order to obtain complete reconstruction, aerial images must be integrated with ground spherical images. In the literature, there are valuable attempts for the integration of aerial-ground images in the context of 3D reconstruction, which can be roughly divided into 2D image-based methods and 3D point cloud-based methods\cite{gao2018accurate}.

\begin{itemize}
	\item \textbf{2D image-based methods.} These methods aim to establish reliable correspondence between aerial and ground images and implement combined bundle adjustment by using both aerial and ground images. The commonly used strategies are image rectification and virtual rendering. The former aims to rectify aerial and ground images onto the same plane in the object space\cite{wu2018integration}; the latter utilizes the rough POS (Positioning and Orientation System) data of ground images and render virtual images from the 3D models reconstructed from aerial images\cite{zhu2020leveraging}. The core idea of these methods is to decrease the geometric distortion caused by large viewpoints. However, for spherical images, their camera imaging model differs from the conventional perspective model. Thus, further consideration should be paid.
	
	\item \textbf{3D point cloud-based methods.} The core idea is to create 3D models separately from aerial and ground images and find reliable enough common 3D correspondences from these 3D models. The integration problem is then converted into the registration of 3D point clouds\cite{gao2018accurate}. However, for spherical images, two issues should be considered. On the one hand, the trajectory of spherical images is limited by the structure of urban streets, which causes large accumulate drift in SfM or SLAM-based image orientation. Thus, it may be hard to model the registration between 3D point clouds by using the commonly used similarity transformation; on the other hand, as shown in Fig.  \ref{fig:figure25}, the reconstructed models of spherical images may contain many false 3D points due to camera distortions. Finding reliable correspondences becomes a non-trivial task between 3D point clouds created from varying images.
\end{itemize}

For the first issue, the recent deep learning-based technique can be used to achieve reliable feature detection and matching. Especially for decreasing the projection distortions, the DCN (deformable convolutional network)\cite{zhu2019deformable} based network can be a promising technique to achieve viewpoint invariant feature detection, and the GCN (graph convolutional network) based network, e.g., SuperGlue\cite{sarlin2020superglue}, can exploit the context information to execute reliable feature matching. For the second issue, the large image orientation problem can be divided into some clusters based on the street structures, and the accumulated drift can be decreased. This divide-and-conquer strategy has been used for efficient and accurate 3D reconstruction of UAV images\cite{jiang2022parallel}.

\section{Conclusions}
\label{sec:5}

Spherical images can record all surrounding environments by using one camera exposure. In contrast to perspective images with limited FOV, spherical images can cover the whole scene and have been increasingly used for 3D modeling in street-view and indoor environments. This paper reviews the 3D reconstruction from spherical images in terms of data acquisition, image matching, image orientation, and dense matching to wrap up recent techniques of 3D modeling based on spherical images. It also presents promising 3D reconstruction applications, including cultural heritage documentation, urban modeling and navigation, tunnel mapping and inspection, and other applications, e.g., urban tree localization, emergency response and rescue, and underwater collision avoidance. Finally, the main prospects are discussed in terms of the exploitation of crowdsource images, the spatial resolution of spherical images, camera distortion and calibration, and integration of aerial-ground images. According to this review, we can conclude that spherical images have great potential in the 3D reconstruction of street view and indoor scenes, and contemporary techniques can support their applications. Future studies can increase the spatial resolution of spherical cameras for data acquisition and exploit current deep learning-based methods to optimize the 3D reconstruction workflow.

\section{Acknowledgment}
\label{sec:6}

The authors would like to thank the anonymous reviewers and editors, whose comments and advice improved the quality of the work. This research was funded by the National Natural Science Foundation of China (Grant No. 42001413) and the Hong Kong Scholars Program (Grant No. 2021-114).

% Can use something like this to put references on a page
% by themselves when using endfloat and the captionsoff option.
\ifCLASSOPTIONcaptionsoff
\newpage
\fi

% trigger a \newpage just before the given reference
% number - used to balance the columns on the last page
% adjust value as needed - may need to be readjusted if
% the document is modified later
%\IEEEtriggeratref{8}
% The "triggered" command can be changed if desired:
%\IEEEtriggercmd{\enlargethispage{-5in}}

% references section

% can use a bibliography generated by BibTeX as a .bbl file
% BibTeX documentation can be easily obtained at:
% http://mirror.ctan.org/biblio/bibtex/contrib/doc/
% The IEEEtran BibTeX style support page is at:
% http://www.michaelshell.org/tex/ieeetran/bibtex/
%\bibliographystyle{IEEEtran}
% argument is your BibTeX string definitions and bibliography database(s)
%\bibliography{mybibfile}
%
% <OR> manually copy in the resultant .bbl file
% set second argument of \begin to the number of references
% (used to reserve space for the reference number labels box)
%\begin{thebibliography}{1}
%
%\bibitem{IEEEhowto:kopka}
%H.~Kopka and P.~W. Daly, \emph{A Guide to \LaTeX}, 3rd~ed.\hskip 1em plus
%  0.5em minus 0.4em\relax Harlow, England: Addison-Wesley, 1999.
%
%\end{thebibliography}

\bibliographystyle{IEEEtran}
\nocite{*}
\bibliography{mybibfile}

\end{document}